\pdfoutput=1
\documentclass[12pt, letterpaper]{article}

\usepackage{times} 
\usepackage{inconsolata} 
\usepackage{indentfirst} 
\usepackage[margin = 1.in]{geometry}  
\usepackage{verbatim} 
\usepackage[T1]{fontenc} 
\usepackage[utf8]{inputenc}
\usepackage[table,usenames,dvipsnames]{xcolor} 
\usepackage{enumitem} 
\usepackage{setspace}
\setlist{topsep=0pt} 
\setlist{itemsep=0pt} 
\setlist{leftmargin=45pt} 
\usepackage{url} 
\usepackage{quoting}

\usepackage{bbm}
\usepackage{amsmath}
\usepackage{bm}
\usepackage{amsfonts}
\usepackage{amssymb}
\usepackage{mathtools}

\usepackage{booktabs}
\usepackage{threeparttable}
\usepackage{adjustbox}
\usepackage{makecell}
\usepackage{tabularx}
\usepackage{multirow}
\usepackage{mdframed}
\usepackage{fancyvrb}
\usepackage[font=small,labelfont=bf]{caption}

\usepackage[authordate, noibid, backend=biber, natbib=true, hyperref=true]{biblatex-chicago}
\DeclareFieldFormat{urldate}{}
\AtEveryBibitem{\clearfield{isbn}}
\AtEveryBibitem{\clearfield{issn}}
\AtEveryBibitem{
  \ifentrytype{misc}{
    \clearfield{eprint}
    \clearfield{primaryclass}
  }{}
}

\DeclareCiteCommand{\parencite}[\mkbibparens]
  {\usebibmacro{prenote}}
  {\usebibmacro{citeindex}%
   \printtext[bibhyperref]{\usebibmacro{cite}}}
  {\multicitedelim}
  {\usebibmacro{postnote}}

\PassOptionsToPackage{breaklinks}{hyperref}
\RequirePackage{hyperref}
\definecolor{darkblue}{rgb}{0,0.08,0.45}
\hypersetup{colorlinks=true, citecolor=darkblue, linkcolor=darkblue, urlcolor=darkblue}

\usepackage{footmisc}

\title{Large Legal Fictions:\\Profiling Legal Hallucinations in Large Language Models\thanks{We thank Neel Guha, Sandy Handan-Nader, Adam T. Kalai, Peter Maldonado, Chris Manning, Joel Niklaus, Kit Rodolfa, Faiz Surani, Andrea Vallebueno, and three anonymous reviewers for helpful discussions and feedback. Our replication materials are available in the Harvard Dataverse (\url{https://doi.org/10.7910/DVN/V4ON8H}) and our data for future benchmarking is available on HuggingFace (\url{https://huggingface.co/datasets/reglab/legal_hallucinations}). Address correspondence to Matthew Dahl, 125 Prospect St., New Haven, CT 06511; email: \url{matthew.dahl@yale.edu.}}}

\author{
    Matthew Dahl,\thanks{J.D. Candidate, Yale Law School, and Ph.D. Student, Yale University Department of Political Science.}\ \
    Varun Magesh,\thanks{Research Fellow, Regulation, Evaluation, and Governance Lab, Stanford University.} \ \
    Mirac Suzgun,\thanks{J.D. Candidate, Stanford Law School, and Ph.D. Candidate, Stanford University Department of Computer Science.} \ \
    Daniel E. Ho\thanks{William Benjamin Scott and Luna M. Scott Professor of Law, Professor of Political Science, Senior Fellow at Stanford Institute for Economic Policy Research, and Faculty Director of the Regulation, Evaluation, and Governance Lab, Stanford University.}
}

\date{April 25, 2024}

\begin{document}
\maketitle

\begin{center}
\Large
\textit{Journal of Legal Analysis (forthcoming)}
\end{center}

\clearpage

\begin{abstract}
\noindent Do large language models (LLMs) know the law? LLMs are increasingly being used to augment legal practice, education, and research, yet their revolutionary potential is threatened by the presence of ``hallucinations''---textual output that is not consistent with legal facts. We present the first systematic evidence of these hallucinations in public-facing LLMs, documenting trends across jurisdictions, courts, time periods, and cases. Using OpenAI’s ChatGPT 4 and other public models, we show that LLMs hallucinate at least 58\% of the time, struggle to predict their own hallucinations, and often uncritically accept users’ incorrect legal assumptions. We conclude by cautioning against the rapid and unsupervised integration of popular LLMs into legal tasks, and we develop a typology of legal hallucinations to guide future research in this area. 
\end{abstract}

\newpage
\doublespacing

\begin{refsection}
\section{Introduction}
\label{sec:introduction}

\begin{figure}[!t]
    \centering\includegraphics[width=4in]{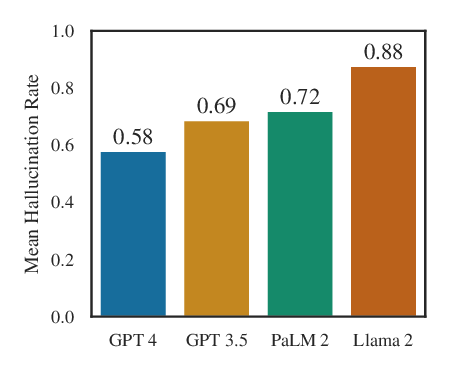}
    \caption{Hallucination rates by LLM, all reference-based tasks pooled. Hallucinations are common across all LLMs when they are asked a direct, verifiable question about a federal court case, but GPT 4 performs best overall.}
    \label{fig:overall}
\end{figure}

How well do large language models (LLMs) know American case law? Modern LLMs such as OpenAI's ChatGPT---tools trained on vast amounts of textual data to predict the next token in a sequence---are driving a transformation in the legal world, from legal education \citep{Choi2024a}, to legal research \citep{Livermore2024}, to legal practice itself \citep{Rodgers2023}. Indeed, recent versions of these artificial intelligence (AI) models seem to excel at law-related tasks, such as first-year law school exams~\citep{Choi2022}, the uniform bar exam~\citep{Katz2023}, statutory reasoning~\citep{Blair-Stanek2023}, and issue-rule-application-conclusion (IRAC) analysis \citep{Guha2023}. But despite the revolutionary potential of these models, a key challenge remains: the issue of ``hallucinations.'' LLMs are liable to generate language that is inconsistent with current legal doctrine and case law, and, in the legal field, where adherence to authorities is paramount, unfaithful or imprecise interpretations of the law can lead to nonsensical---or worse, harmful and inaccurate---legal advice or decisions.

In this work, we present the first evidence documenting the nature, frequency, and correlates of these hallucinations. In doing so, we shed systematic, empirical light on a phenomenon that has so far only received anecdotal treatment in the literature. For example, much media attention has been directed toward a Manhattan lawyer who faced sanctions for using ChatGPT to generate fictional case citations for a brief \citep{Weiser2023}, or another instance where ChatGPT produced a supposed dissent authored by Justice Ruth Bader Ginsburg in the landmark gay rights case \textit{Obergefell v. Hodges}~\citep{Romoser2023}. Even Chief Justice John Roberts, the chief justice of the U.S. Supreme Court, has weighed in on the problem, highlighting hallucinations in his 2023 report on the state of the federal judiciary and arguing that, as yet, ``machines cannot fully replace key actors in court'' \citep[6]{Roberts2023}.

These impressionistic accounts, however, leave unanswered the deeper questions that legal scholars must confront as LLMs continue to grow in popularity. How much legal knowledge is actually embedded in an off-the-shelf LLM? Are LLMs equally familiar with different dimensions of the American common law system---where legal doctrine varies across courts, jurisdictions, and over time---or do they tend to hallucinate more in certain areas than others? When LLMs do hallucinate, do they disproportionately produce false information favoring certain judges or cases? And besides hallucination itself, are there other features of LLMs that legal scholars need to consider---other latent biases or behavioral tendencies that threaten to spill over into downstream applications of these models? Our study seeks to answer these questions, providing insights that are essential for evaluating LLMs' effectiveness in general legal settings.

This research contributes to several literatures. First, there has recently been an explosion of interest in the intersection of law and technology, with a particular focus on the emergence of AI. Much of this work focuses on how lawmakers and administrative agencies ought to govern the deployment of these tools \citep{Engstrom2020, Engstrom2020a, Solow-Niederman2020}, given that they are already being used by public \citep{Engel2021} and private \citep{Barocas2016} actors alike, producing novel privacy concerns \citep{BenShahar2023, King2023} and giving rise to new forms of liability \citep{Henderson2023, Lemley2019, Volokh2023}. As one highly influential but still maturing species of AI, LLMs stand in need of a concrete empirical evaluation of their legal abilities and their legal risks, of which hallucination is certainly one. We supply that information here.

We also contribute to a growing literature regarding the implications of AI for access to justice. Many members of the legal community rightly regard LLMs as a promising solution to the longstanding barriers to adequate legal representation that millions of \textit{pro se} and under-resourced litigants encounter \citep{Chien2024, Perlman2023, Tan2023}. Because they are relatively cheap, easy, and quick to use, LLMs might finally be able to deliver on the federal rules' guarantee of a  ``just, speedy, and inexpensive'' resolution of disputes (Fed. R. Civ. P. 1; \citealp{Roberts2023}). This potential can only be realized, however, if LLMs actually know the law. Additionally, if the legal knowledge embedded in LLMs is not evenly distributed, the widespread adoption of LLMs might unintentionally worsen rather than eliminate current disparities in the availability of legal services \citep{Draper2023, Simshaw2022}. We therefore approach our study of LLMs with an eye toward assessing their ability to truly close the justice gap, examining both their raw hallucination rates as well as any other emergent behaviors that threaten this potential.

Finally, we also contribute to the pressing algorithmic harm literature, which is motivated by the concern that inscrutable algorithms often produce predictions, recommendations, or outputs that are not fairly distributed among individuals or groups \citep{Bar-Gill2023, Gillis2019, Kleinberg2018, Mayson2019}. In our legal setting, the specific danger is that if LLMs do not properly internalize knowledge about some dimension of the law---if LLMs know California law better than Wyoming law, for example, or decisions by Justice Ketanji Brown Jackson worse than decisions by Justice Antonin Scalia, for another---they will regurgitate a falsely homogeneous sense of the legal landscape to their users, collapsing important legal nuances and perpetuating representational harms. Worse, because LLMs are so-called ``foundation'' models, their distributional biases, if they exist, may permeate and afflict \textit{every} downstream version of these models \citep{Bommasani2022}, producing a kind of algorithmic ``monoculture'' by entrenching one particular notion of the law across a wide range of applications \citep{Creel2022, Kleinberg2021}. Accordingly, it is important for legal scholars to obtain a sense of what the correlates of LLMs' hallucinations are, in order to address this new and profound opportunity for cascading algorithmic harms.

Our article proceeds as follows. In Section~\ref{sec:prelims-and-background}, we provide a brief background on LLMs for the non-technical reader and theorize a typology of legal hallucinations. In Section~\ref{sec:tasks}, we develop a set of legal knowledge queries that we use to elicit an LLM's understanding of the law, from simple queries like whether or not a case exists to more complex queries like asking for a statement of a case's holding or its precedential relationship to another case. In Section~\ref{sec:design}, we describe our methodological approach, which entails asking these queries for a random sample of cases across each level of the federal judiciary---the U.S. District Courts (USDC), the U.S. Courts of Appeals (USCOA), and the U.S. Supreme Court (SCOTUS)---and evaluating them using four popular LLMs: OpenAI's ChatGPT 4, OpenAI's ChatGPT 3.5, Google's PaLM 2, and Meta's Llama 2.

In Section~\ref{sec:results}, we present our results. Our findings reveal the widespread occurrence of legal hallucinations: when asked a direct, verifiable question about a randomly selected federal court case, LLMs hallucinate between 58\% (ChatGPT 4) and 88\% (Llama 2) of the time. However, we also find that LLMs perform better on cases that are newer, more salient, and from more prominent legal jurisdictions, suggesting that the risks of legal monoculture are real. We then investigate two additional potential failure points for LLMs, beyond their raw hallucination rates: (1) their susceptibility to contra-factual bias, i.e., their ability to respond to queries anchored in erroneous legal premises \citep{Sharma2023, Wei2023}, and (2) their certainty in their responses, i.e., their self-awareness of their propensity to hallucinate \citep{Azaria2023, Kadavath2022, Tian2023, Xiong2023, Yin2023}. Our results indicate that not only do LLMs often provide seemingly legitimate but incorrect answers to contra-factual legal questions, they also struggle to accurately gauge their own level of certainty without post-hoc recalibration. Accordingly, in Section~\ref{sec:discussion} we conclude that while LLMs appear to offer a way to make legal information and services more accessible and affordable to all, their present shortcomings---particularly in terms of generating accurate and reliable statements of the law---significantly hinder this objective.

\section{Background and Theory}
\label{sec:prelims-and-background}

\subsection{What Is a Language Model?}
\label{sec:problemsetup}

We first provide a brief overview of language models (LMs) for readers who may not necessarily have a deep technical background. LMs can be viewed as functions that map text to text: When a user provides a text input (known as a ``prompt''), the model produces a text output (referred to as a ``response''). If the prompt takes the form of a question, the response can be understood as an answer to that question. An LM generates its response by selecting the most probable sequence of tokens that follow the prompt's tokens; therefore, it essentially functions as a probability distribution over these tokens.

In this work, we focus on \emph{large} language models (LLMs). The largeness of a language model is a dual reference to its parameter count and the scope of its training corpus: LLMs are models that contain billions of parameters and are trained on vast corpora bordering on the size of the Internet. Because of their incredible size, LLMs can be considered general purpose technologies, with the apparent ability to understand and generate human-like text across a wide range of topics, including medicine, finance, education, retail, and law \citep{Eloundou2023}. In contrast to previous forms of machine learning, however, they seem to excel at these tasks despite not being explicitly trained to perform them \citep{Brown2020}; the ``jagged frontier'' of their emergent abilities is still being mapped \citep{DellAcqua2023}.

We also set forth a more formal definition of an LLM, in order to provide the foundation for the typology of legal hallucinations that we develop in the next subsection. We let an LLM be a function $f_{\tau} : \text{prompt} \mapsto \text{response}$, where $f_{\tau}$ operates by sampling responses from a conditional probability distribution that is learned by optimizing over a training corpus hopefully reflective of facts about the world.\footnote{In reality, language generation in LMs actually happens at the level of tokens, not responses themselves; full model responses are constructed autoregressively by sampling $n$ tokens, one at a time, from a distribution $\mbox{Pr}[x_n | x_1, \dots, x_{n-1}]$. We abstract from these details in this article without loss of generality.} We use the symbol $\tau$ to designate a user-configurable ``temperature'' parameter that controls the shape of the probability distribution at inference time. When $\tau=0$, the distribution becomes degenerate and the model's response is theoretically deterministic---the model must always return the most likely response.\footnote{Non-determinism may persist in practice due to a model's implementation details, e.g., the ``mixture of experts'' architecture \citep{Chann2023}.} Following convention, we refer to this deterministic response as the model's ``greedy'' response. As $\tau$ increases, the distribution becomes more uniform and the model's response becomes more stochastic---the model is free to choose from a variety of candidate responses, all of which become more equally likely to be chosen the higher the temperature is. Thus, increasing the temperature of an LLM is one way to potentially increase its hallucination frequency \citep{Lee2023a}. In this article, however, we generally perform our experiments at $\tau=0$, showing that LLMs hallucinate even under the most conservative sampling conditions.

\subsection{The Nature of Legal Hallucinations}
\label{sec:defining_hallucination}

LLMs are showing promise on a number of legal research and analysis tasks \citep{Ash2024, Blair-Stanek2023, Choi2022, Fei2023a, Guha2023, Katz2023, Trozze2023}, but the problem of legal hallucination has so far only been studied in closed-domain applications, such as when a model is used to summarize the content of a given judicial opinion \citep{Deroy2023, Feijo2023} or to synthesize provided legal text \citep{Savelka2023}. In this article, by contrast, we examine hallucination in an open-domain setting, i.e., when a model is tasked with providing an accurate answer to an open-ended legal query. This setting approximates the situation of a lawyer or a \textit{pro se} litigant seeking advice from a legal chat interface.

In the context of such question-answering (QA) scenarios, the study of hallucinations in LMs is still in its infancy, even outside the legal field. There is no universally accepted definition or classification of LM hallucinations \citep{Ji2023, vanDeemter2024, Zhang2023}. However, as \citet{Kalai2023} show, LMs that assign a positive probability to every response token \textit{must} hallucinate at least some of the time. \citet{Xu2024} agree, arguing that ``hallucination is inevitable for any computable LLM, regardless of model architecture, learning algorithms, prompting techniques, or training data.'' Therefore, if hallucinations are here to stay, we believe that it is essential for legal scholars to begin to recognize that there are several different ways in which an LLM can generate false information, as not all modes of hallucination are equally concerning for legal professionals. For example, since hallucinations seem likely to give rise to new forms of tort liability \citep{Henderson2023}, it will be important to differentiate between different types of hallucinations in order to properly assess the predicate elements of such torts. We supply those theoretical resources here, summarizing our typology of legal hallucinations in Table~\ref{tab:HallucinationTypologyTable}. 

First, a model might hallucinate by producing a response that is either unfaithful to or in conflict with the input prompt, a phenomenon canonically referred to as \emph{closed-domain} or \textit{intrinsic} hallucination. This is a major concern in tasks requiring a high degree of accuracy between the response and a long-form input, such as machine translation~\citep{Xu2023} or summarization~\citep{Cao2018}. In legal contexts, such inaccuracies would be particularly problematic in activities like summarizing judicial opinions, synthesizing client intake information, drafting legal documents, or extracting key points from an opposing counsel’s brief.

\begin{table*}

\caption{\label{tab:HallucinationTypologyTable}Typology of legal hallucinations}
\centering
\begin{tabular}[t]{lll}
\toprule
\textbf{Domain} &  \textbf{Type of hallucination} & \textbf{Legal example}\\
\midrule
Closed & response inconsistency with the prompt &  Mischaracterization of an opinion\\
\cmidrule{1-3}
 & response inconsistency with the training corpus & Creative argumentation\\
\cmidrule{2-3}
\multirow{-2}{*}[0.3em]{\raggedright\arraybackslash Open} & response inconsistency with the facts of the world &  Misstatement of the law\\
\bottomrule
\end{tabular}
\end{table*}

Second, an LLM might also hallucinate by producing a response that either contradicts or does not directly derive from its training corpus. Following \citet{Agrawal2023}, we conceptualize this kind of hallucination as one form of \emph{open-domain} or \textit{extrinsic} hallucination. In general, the output of a language model should be logically derivable from the content of its training corpus, regardless of whether the content of the corpus is factually or objectively true.\footnote{For example, if a training corpus consisted of J. K. Rowling's \emph{Harry Potter} series, we would expect an LLM to produce the sentence ``Tom Marvolo Riddle'' in response to a query about Voldemort's real name. However, if the training corpus consisted solely of Jane Austen's \emph{Pride and Prejudice} (for instance), we would consider this LLM output to be a hallucination---because there would be no basis in the training data for making such a claim about Voldemort.} In legal settings, this kind of hallucination poses a special challenge to those aiming to fine-tune the kind of general purpose foundation models that we study in this article with proprietary, in-house work product.\footnote{For example, this kind of firm-specific fine-tuning is the business model of a prominent legal tech startup, Harvey.ai \citep{Ambrogi2023}.} For example, firms might have a catalogue of internal research memos, style guides, and so forth, that they want to ensure is reflected in their bespoke LLM's output. At the same time, however, insofar as \textit{creativity} is valued, certain legal tasks---such as persuasive argumentation---might actually benefit from some lack of strict fidelity to the training corpus; after all, a model that simply parrots exactly the text that it has been trained on could itself be undesirable. As mentioned, creativity can be induced by raising the temperature of the LLM, but responses that are more unpredictable are also those that are more likely to be hallucinations \citep{Lee2023a}. Thus, defining the contours of what counts as an \textit{unwanted} hallucination in this specific sense requires value judgements about the balance between fidelity and spontaneity.

Finally, the third way that an LLM can hallucinate is by producing a response that lacks fidelity to the facts of the world, irrespective of how the LLM is trained or prompted \citep[][]{Maynez2020}. We consider this to be another type of open-domain hallucination, with the key concern being ``factuality'' in relation to the facts of the world~\citep[cf.][]{WittgensteinTractatus}. In our context, this is perhaps the most alarming type of hallucination, as it can undermine the accuracy required in any legal context where a correct statement of the law is necessary.

\subsection{Hallucination Trade-offs} \label{sec:hallucination_trade_offs}

In this article, we investigate only the last kind of hallucination. As mentioned, the first two modes of hallucination are not always problematic in the legal setting: these kinds of hallucinations could actually be somewhat desirable to lawyers if they resulted in generated language that, for example, removed unnecessary information from a given argument (at the expense of being faithful to it) or invented a novel analogy never yet proposed (at the expense of being grounded in the lexicon) \citep{Cao2022}. However, what a lawyer cannot tolerate is the third kind of hallucination, or factual infidelity between an LLM's response and the controlling legal landscape. In a common law system, where \textit{stare decisis} requires attachment to the ``chain'' of historical case law \citep{Dworkin1986}, any misstatement of the binding content of that law would make an LLM quickly lose any professional or analytical utility.

Focusing on non-factual hallucinations alone, however, comes with certain trade-offs. One of the advantages of our typology is that it makes clear that it may not always be possible to minimize all modes of hallucination simultaneously; reducing hallucinations of one kind may increase hallucinations of another. For example, if a given prompt contains information that does not conform to facts about the world, then ensuring response fidelity with respect to the former would by definition produce infidelity---i.e., hallucination---with respect to the latter. More generally, although fidelity to the prompt is necessary for avoiding \textit{closed-domain} hallucination, there is an important sense in which prioritizing such behavior might actually induce the kind of \textit{open-domain} hallucination that we center in this article.

These trade-offs present unavoidable challenges for prospective users of legal LLMs. When responding to a query, should an LLM be skeptical of its prompt or sycophantic to it? If it has been trained on case law from one jurisdiction, should it enforce adherence to that training corpus even when responding about the law in another jurisdiction? If facts about the world conflict with each other---as legal rules often do---should the LLM preserve that nuance or refrain from introducing information outside the scope of a query? Questions like these are ultimately questions about which kinds of legal hallucinations are more and less preferable, and they are questions whose answers require \textit{both} empirical evidence \textit{and} normative arguments. For example, minimizing fact and training corpus hallucinations (at the expense of prompt hallucinations) might be best for avoiding harm to \textit{pro se} litigants, but the calculus might be reversed for sophisticated lawyers who might be less vulnerable to such behavior. We supply some of the empirics that speak to these dilemmas (see Sections~\ref{sec:het_llm_results} and \ref{sec:contrafactual_bias}), but stress that the normative considerations are crucial and should be a topic of continued legal hallucination research.
\section{Profiling Hallucinations Using Legal Knowledge Queries} \label{sec:tasks}

To empirically assess the incidence and correlates of non-factual hallucinations, we adopt a QA framework where the goal is to test an LLM's ability to produce accurate information in response to different kinds of legal queries. We develop fourteen tasks representative of such queries, which we group into three categories in order of increasing complexity and list in Figure~\ref{fig:TaskListTable}.

\begin{table}[t]
    \centering    
    
\begin{tabular}{llll}
\toprule
\textbf{Complexity} & \textbf{Task} & \textbf{Query} & \textbf{Method}\\
\midrule
\multirow{4}{*}{Low} & \multicolumn{1}{l}{\cellcolor{gray!6}{Existence}} & \multicolumn{1}{l}{\cellcolor{gray!6}{Is \texttt{\{case\}} a real case?}} & \multicolumn{1}{l}{\cellcolor{gray!6}{Reference-based}}\\
& \multicolumn{1}{l}{Court} & \multicolumn{1}{l}{What court decided \texttt{\{case\}}?} & \multicolumn{1}{l}{Reference-based}\\
& \multicolumn{1}{l}{\cellcolor{gray!6}{Citation}} & \multicolumn{1}{l}{\cellcolor{gray!6}{What is the citation for \texttt{\{case\}}?}} & \multicolumn{1}{l}{\cellcolor{gray!6}{Reference-based}}\\
& \multicolumn{1}{l}{Author} & \multicolumn{1}{l}{Who wrote the majority opinion in \texttt{\{case\}}?} & \multicolumn{1}{l}{Reference-based}\\
\midrule
\multirow{4}{*}{Moderate} & \multicolumn{1}{l}{\cellcolor{gray!6}{Disposition}} & \multicolumn{1}{l}{\cellcolor{gray!6}{Did \texttt{\{case\}} affirm or reverse?}} & \multicolumn{1}{l}{\cellcolor{gray!6}{Reference-based}}\\
& \multicolumn{1}{l}{Quotation} & \multicolumn{1}{l}{What is a quotation from \texttt{\{case\}}?} & \multicolumn{1}{l}{Reference-based}\\
& \multicolumn{1}{l}{\cellcolor{gray!6}{Authority}} & \multicolumn{1}{l}{\cellcolor{gray!6}{What is an authority cited in \texttt{\{case\}}?}} & \multicolumn{1}{l}{\cellcolor{gray!6}{Reference-based}}\\
& \multicolumn{1}{l}{Overruling year} & \multicolumn{1}{l}{What year was \texttt{\{case\}} overruled?} & \multicolumn{1}{l}{Reference-based}\\
\midrule
\multirow{6}{*}{High} & \multicolumn{1}{l}{\cellcolor{gray!6}{Doctrinal agreement}} & \multicolumn{1}{l}{\cellcolor{gray!6}{Does \texttt{\{case1\}} agree with \texttt{\{case2\}}?}} & \multicolumn{1}{l}{\cellcolor{gray!6}{Reference-based}}\\
& \multicolumn{1}{l}{Factual background} & \multicolumn{1}{l}{What is the factual background of \texttt{\{case\}}?} & \multicolumn{1}{l}{Reference-free}\\
& \multicolumn{1}{l}{\cellcolor{gray!6}{Procedural posture}} & \multicolumn{1}{l}{\cellcolor{gray!6}{What is the procedural posture of \texttt{\{case\}}?}} & \multicolumn{1}{l}{\cellcolor{gray!6}{Reference-free}}\\
& \multicolumn{1}{l}{Subsequent history} & \multicolumn{1}{l}{What is the subsequent history of \texttt{\{case\}}?} & \multicolumn{1}{l}{Reference-free}\\
& \multicolumn{1}{l}{\cellcolor{gray!6}{Core legal question}} & \multicolumn{1}{l}{\cellcolor{gray!6}{What is the core legal question in \texttt{\{case\}}?}} & \multicolumn{1}{l}{\cellcolor{gray!6}{Reference-free}}\\
& \multicolumn{1}{l}{Central holding} & \multicolumn{1}{l}{What is the central holding in \texttt{\{case\}}?} & \multicolumn{1}{l}{Reference-free}\\
\bottomrule
\end{tabular} 

    \caption{Hallucination QA task list. Tasks are sorted in order of increasing complexity. Query wording is paraphrased; see the Online Appendix for exact prompt used. Method column describes the inferential strategy that we use to estimate a hallucination rate for each task: reference-based tasks use known metadata to assess hallucinations, and reference-free tasks use emergent contradictions to assess hallucinations (see Section~\ref{sec:design}).}
    \label{fig:TaskListTable}
\end{table}

\subsection{Low Complexity Tasks}

In the low complexity category, we ask for information that we consider
relatively easy for an LLM to reproduce. The information in this category does not derive from the actual content of a case itself, so it does not require higher-order legal reasoning skills to internalize. Instead, this information is readily available in a case's caption or its syllabus---standard textual locations whose patterns even non-specialized LLMs should be able to recover. We therefore expect LLMs to perform best on these tasks:

\textbf{Existence:}
\textit{Given the name and citation of a case, state whether the case actually exists or not.} This basic evaluation provides preliminary insights into an LLM’s knowledge of actual legal cases: if it cannot distinguish real cases from non-existent ones, it probably cannot offer detailed case insights. We use only real cases in our prompts, so affirming their existence is the correct answer.\footnote{In the Online Appendix, we experiment with using fake cases as well.}

\textbf{Court:}
\textit{Given the name and citation of a case, supply the name of the court that ruled on it.} This task assesses an LLM’s knowledge about legal jurisdictions, an important building block of a case's precedential value. We perform this task across the three different levels of the federal judiciary. Importantly, we note that each level of the judiciary has a different reporter, or the series of volumes that opinions are published in. This is relevant because the reporter is included in the citation that we provide to the LLM, essentially revealing the level of the hierarchy that an opinion is from. All and only SCOTUS cases are published in the \textit{U.S. Reports}. Opinions from the USCOA are published in the \textit{Federal Reporter}, and USDC cases are published in the \textit{Federal Supplement}. Because of this, we expect this task to be more difficult as we descend the hierarchy of courts. There is only one court associated with the U.S. reporter, but 13 associated with the \textit{Federal Reporter}, and 94 associated with the \textit{Federal Supplement}. For USCOA cases, we require the name of the specific circuit court, and for USDC cases, we require the name of the specific district court.

\textbf{Citation:} 
\textit{Given a case name, supply the Bluebook citation of the case.} This query tests an LLM's ability to associate a given dispute with its official record in a reporter volume at a particular page, which is the key way in which different opinions reference and link to each other. For USCOA cases, we further specify that we want the citation for the circuit court opinion, and for USDC cases, we further specify that we want the citation for the district court opinion. We test for citation equality using \textit{eyecite} \citep{Cushman2021}.

\textbf{Author:} 
\textit{Given the name and citation of a case, supply the name of the opinion author.} This query tests an LLM's ability to associate a given case with a particular judge, which is important for contextualizing a case in the broader jurisprudential landscape. For SCOTUS and USCOA cases, we further specify that we want the name of the \textit{majority} opinion author. We accept a fuzzy match of the opinion author's name as accurate.

\subsection{Moderate Complexity Tasks}

Next, in the moderate complexity category, we start to require an LLM to evince knowledge of actual legal opinions themselves. To answer the queries in this category, an LLM must know something about a case's substantive content; these queries seek information that must be collated from idiosyncratic portions of its text. Of course, a database-augmented LLM might still be able to retrieve some of this information without ever actually internalizing the content of a case, but we expect this text-based knowledge to be less available than the information described in the low complexity category. Specifically, we ask for the following information:

\textbf{Disposition:}
\textit{Given a case name and its citation, state whether the court affirmed or reversed the lower court.} This query tests an LLM's knowledge of how the court resolved the instant appeal confronting the parties in the case, which is the first step for determining the holding that is created by the case. Though this is essentially a binary classification task where we accept correct ``affirm'' or ``reverse'' labels as accurate, we consider this task to still be probative of hallucinations because producing the wrong label is still a misstatement of the law. We filter out all ambiguous dispositions (e.g., reversals in part) and we do not ask this query of USDC cases because district courts are courts of original jurisdiction.\footnote{While it is possible for some administrative agency decisions to be appealed to a district court, this occurs infrequently enough that we choose not to ask for case disposition at the district court level.}

\textbf{Quotation:}
\textit{Given a case name and its citation, supply any quotation from the opinion.} This query tests an LLM's ability to produce some portion of an opinion's text verbatim, which is an important feature for lawyers seeking to use a case to stand for a specific proposition. Normally, such memorization is considered an undesirable property of LLMs \citep{Carlini2022}, but in this legal application it is actually desirable behavior. We accept any fuzzy string of characters appearing in the majority opinion as accurate.

\textbf{Authority:}
\textit{Given a case name and its citation, supply a case that is cited in the opinion.} This query probes an LLM's understanding of the chain of precedential authority that supports a given opinion. We do not distinguish between positive and negative citations for this task; we accept any precedent cited in any way in the text of the majority opinion as accurate. We extract and match citations on their volumes, reporters, and pages using \textit{eyecite} \citep{Cushman2021}.

\textbf{Overruling year:}
\textit{Given a case name and its citation, supply the year that it was overruled.} This query tests an LLM's ability to recognize when a given case has been subsequently altered, which is crucial information for lawyer seeking to determine whether a given precedent is still good law or not. This task is the most complicated in this category because it requires the LLM to draw connections between multiple areas of the case space. We accept only the exact year of overruling as accurate, and we limit this task to only those SCOTUS cases that have been explicitly overruled (n=279).\footnote{In Section~\ref{sec:contrafactual_bias}, we experiment with cases that have never been overruled as well.}

\subsection{High Complexity Tasks}

Finally, in the high complexity category, we seek answers to tasks that \textit{both} presuppose legal reasoning skills (unlike the low complexity tasks) \textit{and} are not readily available in existing legal databases like WestLaw or Lexis (unlike the moderate complexity tasks). These tasks all require an LLM to synthesize core legal information out of unstructured legal prose---information that is frequently the topic of deeper legal research. In Section~\ref{sec:zero_resource_design}, we explain how we test LLMs' knowledge of some of these more complex facts without necessarily having access to the ground-truth answers ourselves:

\textbf{Doctrinal agreement:}
\textit{Given two case names and their citations, state whether they agree or disagree with each other.} This query requires an LLM to show knowledge of the precedential relationship between two different cases, information that is essential for higher-order legal reasoning. We use Shepard's treatment codes as a basis for constructing this task, filtering out all ambiguous citation treatments (e.g., neutral treatments) and coarsening the unambiguous codes into ``agree'' and ``disagree'' labels that we accept as accurate. For this task, we use a relatively balanced dataset of 2,839 citing-cited case pairs coded as ``agree,'' and 2,161 citing-cited case pairs coded as ``disagree.'' This task is limited to SCOTUS cases, as our underlying dataset only contains thorough Shepard's data for citations to the Supreme Court.

\textbf{Factual background:}
\textit{Given a case name and its citation, supply its factual background.} This query tests an LLM's understanding of the concrete fact pattern underlying a case, which is helpful in assessing the relevance of the case to current research and in drawing parallels with other cases.

\textbf{Procedural posture:}
\textit{Given a case name and its citation, supply its procedural posture.} This query tests an LLM's understanding of how and why a case has arrived at a particular court, which aids in understanding the precise question presented and standard of review applicable.

\textbf{Subsequent history:}
\textit{Given a case name and its citation, supply its subsequent procedural history, if any.} This query tests an LLM's knowledge of any other related proceedings that concern the given case after a particular decision, which is information that can change or clarify the legal significance of the case.

\textbf{Core legal question:}
\textit{Given a case name and its citation, supply the core legal question at issue.} This query tests an LLM's ability to pinpoint the main issue or issues that a court is addressing in a case, which is the most important factor in assessing whether a case is apposite or not.

\textbf{Central holding:}
\textit{Given a case name and its citation, supply its central holding.} This query tests an LLM's knowledge of the legal principle that a given case stands for, i.e., the precedent that future cases will rely upon or distinguish from. Articulating the holding of a case is crucial for legal analysis and argumentation and is the most complex task that we evaluate.

\section{Experimental Design}
\label{sec:design}

\subsection{Data Construction}
\label{sec:data}

We aim to profile hallucination rates across several legally salient dimensions, including hierarchy, jurisdiction, time, and case prominence. Thus, we construct our test data with an eye toward making statistical inferences on these covariates. 

We begin with the universe of case law from each level of the federal judicial hierarchy---namely, SCOTUS, USCOA, and USDC---that has been published in the volumes of the \emph{U.S. Reports}, the \emph{Federal Reporter}, and the \emph{Federal Supplement}. To ensure balance over time and place,  we then perform stratified random sampling using year strata for the SCOTUS cases, circuit-year strata for the USCOA cases, and state-year strata for the USDC cases. We draw 5,000 cases from each level of the judiciary. Finally, to generate ground-truth answers for our reference-based queries (Section~\ref{sec:reference_based_querying}), we merge these units with metadata obtained from the \citet{CaselawAccessProject2023}, the Supreme Court Database \citep{Spaeth2022}, the Appeals Courts Database Project \citep{Songer2008, Kuersten2011}, the Library of Congress \citep{Congress.gov2023}, and Shepard's Citations \citep{Fowler2007, Black2013}.\footnote{More information about how we use these metadata to construct each query is available in the Online Appendix.}

\subsection{Reference-based Querying}
\label{sec:reference_based_querying}

The most straightforward way to study hallucinations in the open-domain setting is to use a test oracle---or an external \emph{reference}---to detect and adjudge non-factual responses \citep{Lin2022, Lee2023, Li2023}. Such oracles are usually difficult and costly to construct \citep{Krishna2021}, but we use the tabular metadata described in Section~\ref{sec:data} to develop ours. Our design exploits the fact that while LLMs are known to have been trained on the raw text of American case law, which is in the public domain \citep{Henderson2022}, they have likely \textit{not} been trained on these cases' attendant metadata, which exist separately from the cases' textual content and which we have aggregated from disparate sources.

These metadata enable us to construct reference-based queries for the first nine of our tasks (Figure~\ref{fig:TaskListTable}). These queries take the form of $N$ question-and-answer triples (prompt, response, $\text{response}^\prime$), where prompt is a case-specific question, response is the LLM's greedy answer retrieved from calling $f_0(\cdot)$, and $\text{response}^\prime$ is the known ground-truth answer.\footnote{Recall from Section~\ref{sec:problemsetup} that the $f_0$ notation represents performing inference with the LLM at temperature zero---i.e., under its deterministic behavior.} Our estimand of interest for each task is the population-level hallucination rate $\pi$, which we estimate by averaging over the $N$ sampled queries:

\begin{equation}
\label{eq:pi_hat_resource_aware}
	\pi = \hat{\pi} = \frac{1}{N} \sum \mathbbm{1}[\text{response} \neq \text{response}^\prime]
\end{equation}

Occasionally, an LLM will produce a response that is neither a hallucination nor a correct answer, but rather an explicit \textit{abstention} from answering the question. For example, the LLM might admit that it does not know the answer or demur that it is unable to provide the answer for some reason, perhaps due to safety concerns. In these instances, we nevertheless count the response as a non-hallucination, on the theory that an LLM cannot hallucinate when it is affirmatively abstaining from responding \citep{Feng2024}. We document the frequency of these abstentions in the Online Appendix, but they are generally rare and do not substantively affect our findings.

\subsection{Reference-free Querying} \label{sec:zero_resource_design}

Reference-based querying lets us directly recover our population parameter of interest, but two problems limit the effectiveness of the approach. First, we are restricted to asking questions for which digestible metadata exist and a clear answer has been recorded, which rules out many more complex inquires. Second, precisely because these queries can be answered with tabular data, legal database-augmented LLMs \citep{Cui2023, Savelka2023} are likely to soon solve or at least mask hallucinated responses to these queries \citep{Peng2023, Shuster2021}.

To test the tasks that cannot be easily verified against an external legal database, we employ reference-free querying instead, which detects hallucinations by exploiting the stochastic behavior of LLMs at higher temperatures \citep{Agrawal2023, Manakul2023, Min2023}. This approach is rooted in the theory that hallucinations are more likely to originate in flat probability distributions with higher next-token uncertainties, whereas factual answers should always have a high probability of being the generated response given a prompt. Thus, by repeatedly querying an LLM at a non-greedy temperature, we can estimate the model's hallucination rate by examining its self-consistency---factual responses should not change, but hallucinated ones will.

Most reference-free approaches implicitly assume that the LLM is calibrated, i.e., that there is indeed some correlation between its self-consistency and its propensity to hallucinate. For reasons that we discuss in Section~\ref{sec:calibration}, we are unwilling to make this assumption in our legal setting. We therefore adopt a slightly different implementation that is still reference-free, but only requires \textit{contradiction}, not \textit{consistency} \citep{Mundler2023}. Specifically, for our final five tasks (Figure~\ref{fig:TaskListTable}), we construct reference-free queries in the form of $N$ question-and-answer triples (prompt, $\text{response}^{(1)}$, $\text{response}^{(2)}$), where prompt is the question, $\text{response}^{(1)}$ is one LLM answer retrieved by calling $f_1(\cdot)$ once, and $\text{response}^{(2)}$ is another LLM answer retrieved by calling $f_1(\cdot)$ again.\footnote{Recall from Section~\ref{sec:problemsetup} that the $f_1$ notation represents performing inference with the LLM at temperature one---i.e., with some degree of stochasticity.} Detecting a hallucination then amounts to detecting a logical contradiction between the two stochastic answers: any such contradiction guarantees non-factuality, because two contradictory answers cannot both be correct.

To identify these contradictions at scale, we feed both answers into GPT 4 and ask it for its assessment. This technique does not assume anything about $f_1(\cdot)$'s calibration---it just requires that GPT 4 possess logical reasoning skills sufficient to compare $f_1(\cdot)$'s two responses and accurately label them as contradictory as not. To justify this reliance on GPT 4, we manually label a portion of the reference-free responses ourselves and conduct an intercoder reliability analysis to ensure that GPT 4 is indeed able to perform this task. Full information about our procedure and a validity check is provided in the Online Appendix (We find that GPT 4's reliability is comparable to human labeling of contradictions.)

An important caveat of this approach is that it only allows us to establish a \textit{lower bound} on the hallucination rate for our reference-free queries:

\begin{equation}
	\pi \geq \hat{\pi} = \frac{1}{N} \sum \mathbbm{1}[\text{response}^{(1)} \neq \text{response}^{(2)}]
\end{equation}

\noindent 
Although self-contradiction guarantees hallucination, the inverse does not hold: two answers may be logically consonant but still lack fidelity to the law. Because we are unwilling to assume calibration, we accept this inferential limitation, but, as we show below, even the lower bounds on hallucination rates are quite high and informative.

\subsection{Models}

We perform our experiments using four popular, state-of-the-art, off-the-shelf LLMs: 
\begin{enumerate}
    \item OpenAI's ChatGPT 4~\citep[\texttt{gpt-4-1106-preview},][]{OpenAI2023},
    \item OpenAI’s ChatGPT 3.5~\citep[\texttt{gpt-3.5-turbo-0613},][]{OpenAI2023a},
    \item Google's PaLM 2~\citep[\texttt{text-bison-001},][]{Anil2023}, and
    \item Meta's Llama 2~\citep[\texttt{Llama-2-13b-chat-hf},][]{Touvron2023}.
\end{enumerate}

We run each query under both ``zero-shot" and ``three-shot" prompting setups. In the zero-shot setup, we simply ask the LLM about the given case directly, whereas in the three-shot setup, we prepend several example questions and responses to give the LLM an opportunity to perform in-context learning \citep{Brown2020}. We provide the full text of the prompts we use for each query, along with the few-shot examples, in the Online Appendix. In total, we execute more than 800,000 queries---200,000+ per LLM---and we share our raw API calls and model responses in the replication materials accompanying this article.
\section{Results} 
\label{sec:results}

We begin by presenting our main results profiling LLMs' hallucination rates, which cut to the core of popular concerns over LLMs' suitability for legal applications (Section~\ref{sec:hallucination_rate_results}). Then, after showing that hallucinations are generally widespread, and highlighting the correlates of LLMs' hallucination rates, we turn to two additional challenges that threaten LLMs' utility for legal adoption: (1) their susceptibility to contra-factual bias, i.e., their ability to handle queries based on mistaken legal premises (Section~\ref{sec:contrafactual_bias}), and (2) their certainty in their responses, i.e., their self-awareness of their propensity to hallucinate (Section~\ref{sec:calibration}).

\subsection{Hallucination Rates and Heterogeneity} \label{sec:hallucination_rate_results}

Tables~\ref{tab:CategoryOneTable}, \ref{tab:CategoryTwoTable}, and \ref{tab:CategoryThreeTable} report our estimated hallucination rates and their standard errors for each category of our tasks. We find that hallucinations vary with the substantive complexity of the task (Section~\ref{sec:het_substance_results}), the hierarchical level of the court (Section~\ref{sec:het_hierarchy_results}), the jurisdictional location of the court (Section~\ref{sec:het_jurisdiction_results}), the prominence of the case (Section~\ref{sec:het_prominence_results}), the year the case was decided (Section~\ref{sec:het_time_results}), and the LLM queried (Section~\ref{sec:het_llm_results}). We do not find substantial differences between zero-shot and few-shot prompting, so we focus our discussion on the few-shot results alone.

\subsubsection{Hallucinations Vary by Task Complexity}
\label{sec:het_substance_results}

\begin{figure}[!t]
	\centering\includegraphics[width=4in]{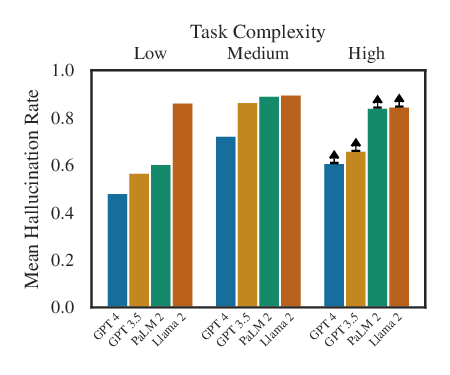}
	\caption{
		Relationship between task complexity and mean hallucination rate. Higher values indicate a greater likelihood of factually incorrect LLM responses. High complexity tasks include several reference-free tasks, so those reported hallucination rates are lower bounds on the true rates. Contra-factual tasks and the doctrinal agreement high complexity task are excluded from this comparison.
	}
	\label{fig:complexity}
\end{figure}

As we hypothesized in Section~\ref{sec:tasks}, we first observe that hallucinations increase with the complexity of the legal research task at issue, which we visualize in Figure~\ref{fig:complexity}. Starting with the low complexity category (Table~\ref{tab:CategoryOneTable}), the LLMs perform best on the simple \textbf{Existence} task, though this is in part driven by their tendency to always answer ``yes'' when asked about the existence of \textit{any} case. (In the Online Appendix we demonstrate this problem by asking about the existence of fake cases instead.) The models begin to struggle more when prompted for information about a case's \textbf{Court}, \textbf{Citation}, or \textbf{Author}. Hallucinations then surge among the moderate complexity tasks (Table~\ref{tab:CategoryTwoTable}), all of which require the LLMs to evince knowledge of the actual content of a legal opinion. We note that these results are not just a product of different evaluation metrics: although the \textbf{Quotation} task, for example, requires near-word reproduction of particular sentences and phrases to be judged correctly, the \textbf{Disposition} task simply asks for binary responses from the model. Yet, the LLMs hallucinate widely in both setups.

The results for the high complexity tasks (Table~\ref{tab:CategoryThreeTable}) confirm this general pattern of poor performance. Starting with \textbf{Doctrinal agreement}, recall that this query asks the LLM to make an analogical judgment about the precedential relationship between two given cases, for which we have ground-truth labels from Shepard's treatment codes. Because this is another binary classification task, the LLMs' hallucination rates on this task---near 0.5---represent little improvement over random guessing, and are actually sometimes worse. This suggests that LLMs know little about substantive legal doctrine, calling into question their ability to accurately assist lawyers in more realistic, applied settings.

The remaining tasks in the high complexity category amplify these concerns, but it is important to keep in mind that the hallucination rates that we report for these tasks are only \textit{lower bounds} on the true rates, as these tasks are evaluated using our reference-free method (Section~\ref{sec:zero_resource_design}). To provide some context for these bounds, we note that in a similar self-contradiction setup, \citet{Mundler2023} found that GPT 3.5 hallucinated about 14.3\% of the time on general QA queries. On our legal QA queries, GPT 3.5 and our other LLMs far surpass this baseline rate---and it is possible that the true hallucination rate is even higher.

For example, we find that even on the easier reference-free tasks---\textbf{Factual background} and \textbf{Procedural posture}---our LLMs hallucinate at least 49\% of the time. Performance degrades further on the most complex \textbf{Core legal question} and \textbf{Central holding} tasks, with hallucinations arising in response to at least 59\% and 63\% of our queries, respectively. Hallucinations are lowest among GPT 4 responses to the \textbf{Subsequent history} task at the SCOTUS level, but this is because the model simply tends to state that the litigation concluded with the Supreme Court decision. This may not actually be correct---many Supreme Court cases result in a remand and have additional procedural history in lower courts. However, we are unable to capture this kind of mistake, as our methodology only permits us to identify hallucinations where the model contradicts itself. We are not able to capture repeated incorrect answers as instances of hallucination, meaning that our estimate of hallucination in the SCOTUS \textbf{Subsequent history} task is likely to understate the rate of hallucination by a larger margin that other tasks.

Taken together, these results invite skepticism about LLMs' true knowledge of the law. Our reference-free tasks, in particular, raise serious doubts about LLMs' knowledge of substantive aspects of American case law---the very knowledge that attorneys must often synthesize themselves, instead of merely looking up in a database.

\begin{figure}[!t]
    \centering\includegraphics[width=4in]{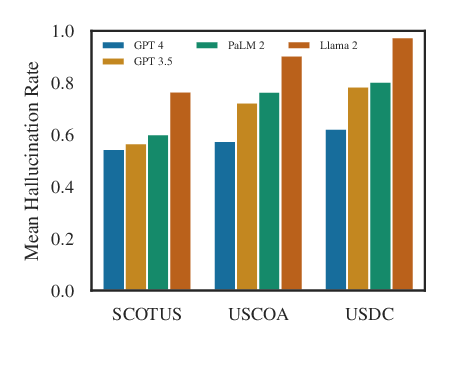}
	\caption{
		Relationship between judicial hierarchy and mean hallucination rate,
		all reference-based tasks pooled. 
        Hallucination rates are higher for lower levels of the federal judiciary.
	}
    \label{fig:court_level}
\end{figure}

\subsubsection{Hallucinations Vary by Court}
\label{sec:het_hierarchy_results}

We next examine trends by hierarchy, exploring LLMs' abilities to restate the case law of the three different levels of the federal judiciary. We find that across all tasks and all LLMs, hallucinations are lowest in the highest levels of the judiciary, and vice-versa (Figure \ref{fig:court_level}). Thus, our LLMs perform best on tasks at the SCOTUS level, worse on tasks at the USCOA level, and worst on tasks at the USDC level. These results are encouraging insofar as it is important for LLMs to be knowledgeable about the most authoritative and wide-ranging precedents, but discouraging insofar as they suggest that LLMs are not well attuned to localized legal knowledge. After all, the vast majority of litigants do not appear before the Supreme Court, and may benefit more from knowledge that is tailored to their home district court---their court of first appearance.

\subsubsection{Hallucinations Vary by Jurisdiction}
\label{sec:het_jurisdiction_results}

\begin{figure}[!t]
    \centering
	\includegraphics[width=5in]{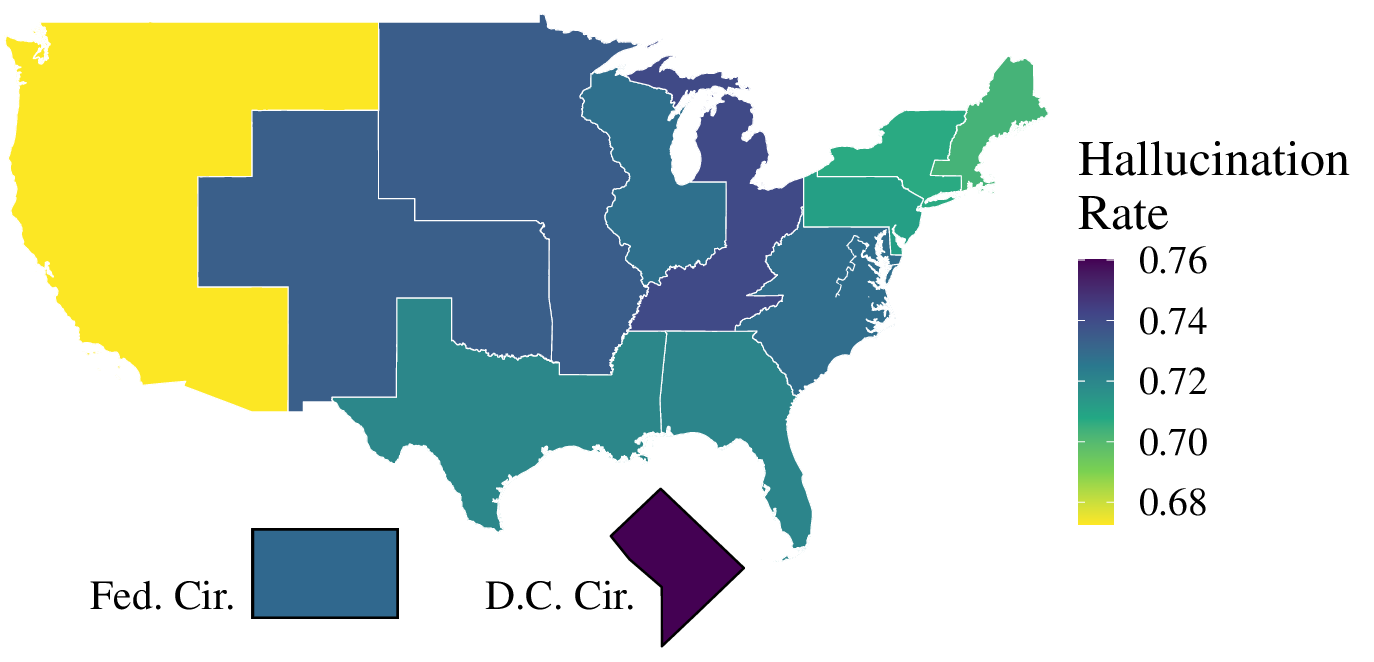}
    \vspace{.5em}
	\caption{
		Relationship between USCOA jurisdiction and mean hallucination rate, all reference-based USCOA tasks and models pooled, post-1981 cases only. LLM performance is strongest in jurisdictions that are commonly perceived to play a more influential role.
	}
    \label{fig:court_coa_geography_since_1981}
\end{figure}

To better understand the relationship between different courts and hallucinations, we next zoom in on the middle level of the judicial hierarchy---the Courts of Appeals---and examine horizontal heterogeneity across the circuits.\footnote{Because not all Courts of Appeals were created at the same time, for parity in comparison here we exclude from our results cases decided before 1982, the year the youngest circuit---the Federal Circuit---was created. We report the full, non-truncated results in the Online Appendix, which are largely consistent with these post-1981 results.} Figure~\ref{fig:court_coa_geography_since_1981} depicts these results geographically, showing lower hallucination rates in lighter colors and higher rates in darker colors. Pooling our tasks and models together, we see the best performance in the Ninth Circuit (comprising California and adjacent states in yellow), the Second Circuit (comprising New York and adjacent states in soft green), the Third Circuit (comprising Pennsylvania and adjacent states in soft green), and the First Circuit (comprising Maine and adjacent states in soft green). By contrast, performance tends to be worst in the circuits in the geographic center of the country.

These results confirm popular intuitions about the influential role that the Second, Third, and Ninth Circuits play in the American legal system. Because it encompasses New York City, the Second Circuit has traditionally had a significant impact on financial and corporate law, and many landmark decisions in securities law, antitrust, and business litigation have come from this court. The Third Circuit enjoys similar influence in the corporate law domain owing to Delaware's status as the legal home for many corporations. Finally, the Ninth Circuit handles more cases than any other federal appellate court, and often issues rulings that advance progressive positions that lead to disproportionate review by the Supreme Court.

Perhaps surprisingly, however, our results stand in tension with received wisdom about the D.C. Circuit, which is generally thought to be the \textit{most} influential appellate division. In our tasks, our LLMs actually perform worst on this circuit. This counterintuitive finding is one example of the way that unanticipated biases might trouble the reliance on LLMs in practice. 

\subsubsection{Hallucinations Vary by Case Prominence}
\label{sec:het_prominence_results}

\begin{figure}[!t]
    \centering
    \includegraphics[width=4in]{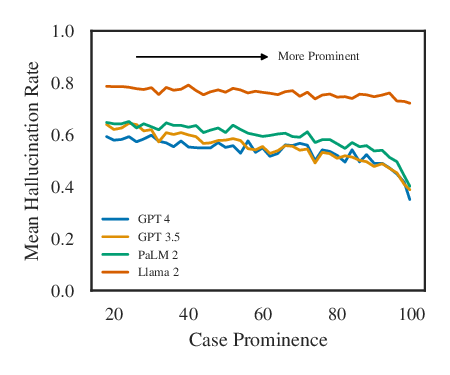}
	\caption{
		Relationship between SCOTUS case prominence (measured by PageRank percentile) and mean hallucination rate, all SCOTUS tasks pooled. Hallucinations decline sharply as case prominence passes the 90th percentile, meaning that LLMs are more likely to respond with accurate information about prominent cases.
	}
    \label{fig:importance}
\end{figure}

To probe the role of legal prominence more directly, we move to SCOTUS-level results next, examining the relationship between \textit{case} importance and hallucinations. To measure case prominence within this single level of the judiciary, we use the Caselaw Access Project's PageRank percentile scores, a metric of citation network centrality that captures the general legal and political prominence of a case.

We find that case prominence is negatively correlated with hallucination, reaffirming our results from above (Figure~\ref{fig:importance}). However, we also note that a sharp slope change occurs around the 90th prominence percentile in the GPT 4, GPT 3.5, and PaLM 2 models. This suggests that the bias of these LLMs---but not Llama 2---may be skewed even more toward the most well-known decisions of the American legal system, even within the SCOTUS level.

\subsubsection{Hallucinations Vary by Case Year}
\label{sec:het_time_results}

\begin{figure}[!t]
    \centering
    \includegraphics[width=4in]{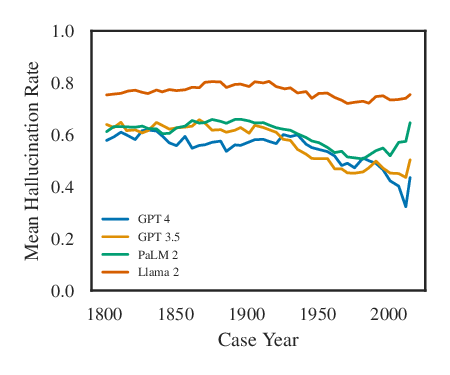}
	\caption{
		Relationship between SCOTUS case decision year and mean hallucination rate, all SCOTUS tasks pooled. LLMs are most likely to respond with accurate information in cases from the latter half of the 20th century, struggling on very old and very new cases.
	}
    \label{fig:year}
\end{figure}

Because case law develops in virtue of new decisions building on old ones over time, the age of a case may be another useful predictor of hallucination. Examining this relationship at the SCOTUS level in Figure~\ref{fig:year}, we find a non-linear correlation between hallucination and age: hallucinations are most common among the Supreme Court's oldest and newest cases, and least common among its post-war Warren Court cases (1953-1969). This result suggests another important limitation on LLMs' legal knowledge that users should be aware of: LLMs' peak performance may lag several years behind the current state of the doctrine, and LLMs may fail to internalize case law that is very old but still applicable and relevant law.

\subsubsection{Hallucinations Vary by LLM}
\label{sec:het_llm_results}

Finally, we also partition our results by the LLM itself and compare across models. We find that not all LLMs are equal: as expected, GPT 4 performs best overall, followed by GPT 3.5, followed by PaLM 2, followed by Llama 2 (Figure \ref{fig:overall}).

\begin{figure}[!t]
    \centering
	\includegraphics[width=4in]{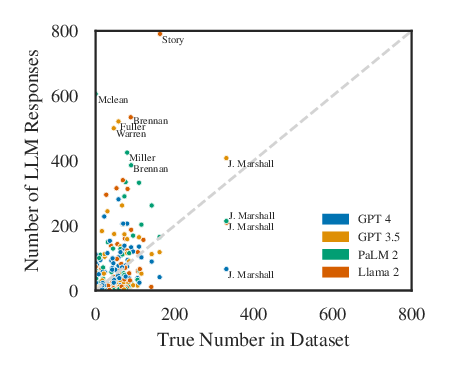}
	\caption{Number of times each justice is stated to be the author of a SCOTUS case versus the actual number of cases authored by each justice in our time period-stratified dataset. A small number of justices are disproportionately represented in LLM responses.}
    \label{fig:author_inductive_bias}
\end{figure}

We also discover tendencies towards different inductive biases, or the predisposition of an LLM to generate certain outputs more frequently than others. In Figure \ref{fig:author_inductive_bias}, we highlight one of these biases for our SCOTUS-level \textbf{Author} task, which asks the LLM to supply the name of the justice who authored the majority opinion in the given case. Each LLM we test has slightly different inductive preferences; some err towards the most recognizable justices, but others are a little more difficult to explain. For example, Llama 2 disproportionately favors Justice Story---an influential jurist who authored the famous \textit{Amistad} opinion, among others---whereas PaLM 2 prefers Justice McLean---also an important jurist, but one more known for his dissents than his majority opinions, such as his dissent in the infamous \textit{Dred Scott} case. Across the board, all our LLMs tend to overstate the true prevalence of justices at a higher magnitude than they understate them, as indicated by the greater dispersion of the points above the $y=x$ line in Figure~\ref{fig:author_inductive_bias}.

These biases demonstrate one way that LLMs inevitably encounter the kind of hallucination trade-off that we discuss in Section~\ref{sec:hallucination_trade_offs}. If the inductive bias that an LLM learns from its training corpus is not well-aligned with the true distribution of facts about the world, then the LLM is likely to make systematic errors when queried about those facts. Moreover, the persistence of inductive biases also increases the risk of LLMs instantiating a kind of legal monoculture \citep{Kleinberg2021}. Instead of accurately restating the full variation of the law, LLMs may simply regurgitate information from a few prominent members of the response set that they have been trained on, flattening legal nuance and producing a falsely homogenous sense of the legal landscape.

\subsection{Contra-factual Bias} \label{sec:contrafactual_bias}

We now turn to the first of two potential failure points that we seek to examine for LLMs performing legal tasks, beyond their sheer propensity to hallucinate: their bias toward accepting legal premises that are not anchored in reality and answering queries accordingly. We view this behavior as a particular kind of model sycophancy (the tendency of an LLM to agree with a user's preferences or beliefs, even when the LLM would reject the belief as wrong without the user's prompting; \citealp{Sharma2023, Wei2023}) or general cognitive error \citep{Tversky1974, Jones2022, Suri2023}.

This bias poses a subtle but pernicious challenge to those aiming to use LLMs for legal research. When a researcher is learning about a topic, they are not only unsure about the \textit{answer}, they are also often unsure about the \textit{question} they are asking as well. Worse, they might not even be aware of any defects in their query; research by its nature ventures into the realm of ``unknown unknowns'' \citep{Luft1955}. This is especially true for unsophisticated \textit{pro se} litigants, or those without much legal training to begin with. Relying on an LLM for legal research, they might inadvertently submit a question premised on non-factual legal information or folk wisdom about the law. As discussed in Section~\ref{sec:hallucination_trade_offs}, this then forces a trade-off: if the LLM is too intent on minimizing prompt hallucinations, it runs the risk of simply accepting the user's misconception as true and producing a factual hallucination instead.

\begin{figure}[t]
    \centering
    \includegraphics[width=4in]{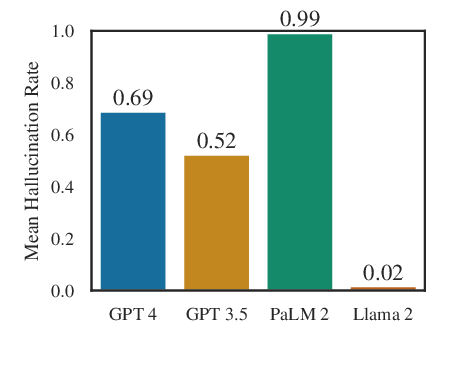}
    \caption{Hallucination rates by LLM, all contra-factual tasks pooled. Llama 2 is very unlikely to hallucinate on these tasks because it almost always rejects the premise in the question. However, this tendency also leads it to perform more poorly on tasks with correct premises (cf. Figure~\ref{fig:overall}).}
    \label{fig:false_premise}
\end{figure}

To test whether this risk is real in the legal setting, we evaluate two modified versions of our reference-based queries, but with premises that are false by construction. Specifically, we ask the LLMs to (1) provide information about an author's dissenting opinion in an appellate case in which they did not in fact dissent and (2) furnish the year that a SCOTUS case that has never been overruled was overruled. In both cases, we consider failing to provide the requested information an acceptable answer; any uncritical answering of the prompt is treated as a hallucination.

Table~\ref{tab:ContraryToFactTable} reports the results of this experiment and Figure~\ref{fig:false_premise} summarizes them by LLM. In general, LLMs seem to suffer from contra-factual bias on these legal information tasks. As in the raw hallucination tasks, contra-factual bias hallucinations are higher in lower levels of the judiciary. Substantively, they are also greatest for the question with a false overruling premise, possibly reflecting the increased complexity of the question asked.

Llama 2 performs exceptionally well, demonstrating little contra-factual hallucination. However, this success is linked to a different kind of hallucination---in many false dissent examples, for instance, Llama 2 often states that the case or justice does not exist at all. (In reality, all of our false dissent examples were created with real cases and real justices---just justices who did not author a dissent for the case.) Under our metrics for contra-factual hallucination, we choose to record these examples as successful rejections of the premise. The kind of error that Llama 2 makes here is already measured in its poor performance on other tasks, especially \textbf{Existence}.

\subsection{Model Calibration} \label{sec:calibration}

The second potential hazard that we investigate is model calibration, or the ability of LLMs to ``know what they know.'' Ideally, a well-calibrated model would be confident in its factual responses, and not confident in its hallucinated ones \citep{Azaria2023, Kadavath2022, Tian2023, Xiong2023, Yin2023}. If this property held for legal queries, users would be able to adjust their expectations accordingly and could theoretically learn to trust the LLM when it is confident, and learn to be more skeptical when it is not \citep{Zhang2020}. Even more importantly, if an LLM knew when it was likely to be hallucinating, the hallucination problem could be in principle solvable through some form of reinforcement learning from human feedback (RLHF) or fine-tuning, with unconfident answers simply being suppressed \citep{Tian2023a}.

\begin{figure}[!t]
    \centering
\includegraphics[width=4in]{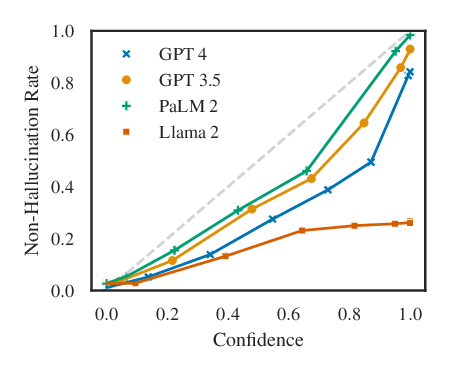}
    \caption{Calibration curves by LLM, all reference-based tasks pooled. PaLM 2 is best calibrated on legal queries, followed by GPT 3.5, GPT 4, and lastly Llama 2, which is significantly worse than the first three models.}
    \label{fig:ece}
\end{figure}

To study our LLMs' calibration on legal queries, we estimate the expected calibration error (ECE) for each of our tasks. We describe our estimation strategy in full in the Online Appendix, but, intuitively, it entails extracting a confidence score for each LLM answer that we obtain and comparing it to the empirical hallucination rate that we observe. Table~\ref{tab:ECETable} reports the results of this analysis at the task level, and Figure~\ref{fig:ece} pools our findings at the LLM level by plotting those two metrics---confidences and empirical non-hallucination frequencies---against each other, binned into 10 equally-sized bins (represented by the dots). In a perfectly calibrated model, the confidences and empirical frequencies would be perfectly correlated along the $y=x$ diagonal.

Overall, we note that PaLM 2 ($\text{pooled ECE} = 0.057$), GPT 3.5 ($\text{pooled ECE} = 0.099$), and GPT 4 ($\text{pooled ECE} = 0.190$) are significantly better calibrated than Llama 2 ($\text{pooled ECE} = 0.421$). Interestingly, although GPT 4 is our best performing model in terms of raw hallucination rates (Figure~\ref{fig:overall}), it is actually less calibrated than PaLM 2 and GPT 3.5, which are otherwise inferior. This suggests that even the newest and most advanced LLMs may not always be superior in all desirable senses---although GPT 4 is currently the LLM least prone to hallucination, our results imply that when it \textit{does} hallucinate, it does so in a way that is more likely to mislead users than GPT 3.5 or PaLM 2.

Diving into the task-level results (Table~\ref{tab:ECETable}), we see that across all LLMs, calibration is poorer on our more complex tasks, like \textbf{Doctrinal agreement}, and on tasks directed toward lower levels of the judicial hierarchy. ECE is also higher on our partially open-ended tasks such as \textbf{Court} and \textbf{Author}. In these tasks, the LLM has a large but finite universe of responses, and the high ECE for these tasks reflects the LLMs' tendencies to over-report on the most prominent or widely known members of the response set.

In all cases, the calibration error is in the positive direction: our LLMs systematically \textit{overestimate} their confidence relative to their actual rate of hallucination.\footnote{In the Online Appendix, we explore whether this bias can be corrected with an \textit{ex post} scaling adjustment, but conclude that challenges remain.} This finding, too, suggests that users should exercise caution when interpreting LLMs' responses to legal queries, especially those of Llama 2. Not only may they receive a hallucinated response, but they may receive one that the LLM is overconfident in and liable to repeat again.
\section{Discussion}
\label{sec:discussion}

We began this article with a question that has surged in salience over the last twelve months: Will AI systems like ChatGPT soon reshape the practice of law and democratize access to justice? Although there is much enthusiasm for LLMs' potential to revolutionize these domains, we highlight the problem of legal hallucinations, which remains a serious obstacle to the adoption of these models. Performing the first systematic empirical test of popular perceptions \citep{Roberts2023, Romoser2023, Weiser2023}, we show that factual legal hallucinations are widespread in the LLMs that we study---OpenAI's ChatGPT 4, OpenAI's ChatGPT 3.5, Google's PaLM 2, and Meta's Llama 2---on the bulk of the legal knowledge tasks that we profile (Section~\ref{sec:hallucination_rate_results}).

We also push beyond conventional wisdom by documenting the correlates of these hallucinations and by surfacing two additional behaviors that threaten LLMs' utility for legal applications: (1) their susceptibility to contra-factual bias, i.e., their inability to handle queries containing an erroneous or mistaken starting point (Section~\ref{sec:contrafactual_bias}), and (2) their certainty in their responses, i.e., their inability to always ``know what they know'' (Section~\ref{sec:calibration}). Unfortunately, we find that LLMs frequently provide seemingly genuine answers to legal questions whose premises are false by construction, and that under their default configurations they are imperfect predictors of their own tendency to confidently hallucinate legal falsehoods.

These findings complicate the existing literature that suggests that LLMs are performing increasingly well on a number of legal benchmarking tasks \citep{Ash2024, Blair-Stanek2023, Choi2022, Fei2023a, Guha2023, Nay2023, Katz2023, Trozze2023}. Our study is related to this prior research, but is oriented in a slightly different vein. Instead of examining LLMs' ability to engage in legal \textit{reasoning}, we assess LLMs' capacity to internalize legal \textit{knowledge}. Ultimately, LLMs will need to excel in both of these respects if they are going to be effectively integrated into the legal profession. So long as they suffer from gaps in their background legal knowledge---as our results suggest---they will be unable to function as reliable sources of legal counsel and advice, no matter how strong their in-context reasoning abilities become.

Our results therefore temper optimism for the ability of off-the-shelf, publicly available LLMs to accelerate access to justice \citep{Perlman2023, Tan2023, Tito2017}. Indeed, our findings suggest that the risks of using these generic foundation models are especially high for litigants who are:

\begin{enumerate}
    \item Filing in courts lower in the judicial hierarchy or those located in less prominent jurisdictions,
    \item Seeking more complex forms of legal information,
    \item Formulating questions with mistaken premises, or
    \item Unsure of how much to trust the LLMs' responses.
\end{enumerate}

In short, we find that the risks are \textit{highest} for those who would benefit from LLMs \textit{most}---under-resourced or \textit{pro se} litigants. Some of these risks---namely, (3) and (4)---might be mitigated with improved user education, but others---(1) and (2)---are more intractable. LLMs would ideally do best at localized legal information (rather than SCOTUS-level information), be able to correct users when they ask misguided questions (rather than accepting their premises at face value), and be able to moderate their responses with the appropriate level of confidence (rather than hallucinating with conviction). Consquently, we echo concerns that the proliferation of LLMs may ultimately exacerbate, rather than eradicate, existing inequalities in access to legal services \citep{Draper2023, Simshaw2022}. At the same time, increased reliance on LLMs also has the potential to produce a kind of legal monoculture \citep{Creel2022, Kleinberg2021}, with users being fed information from only a limited subset of judicial sources that elide many of the deeper nuances of the law. This new monoculture, in turn, is likely to reify the geographic, temporal, and judge-level biases that we diagnose above, as the foundation-like property of these models permits those biases to propagate into any downstream tools built on top of the original LLM \citep{Bommasani2022}.

Some recent research suggests that hallucinations can be diminished through the adoption of techniques like retrieval-augmented generation (RAG) \citep{Shuster2021, Cui2023, Peng2023, Savelka2023}, advanced prompting (such as chain-of-thought prompting or chain-of-verification)~\citep{si2023prompting, lei2023chain, Mundler2023, ji-etal-2023-towards, dhuliawala2023chain, suzgun2024meta}, specialized fine-tuning~\citep{Tian2023a, razumovskaia2023dial, zhang2023r}, factuality-focused decoding methods~\citep{shi2023trusting, mallen2023not, li2024inference, chuang2024dola}, or external database checks~\citep{chern2023factool, Peng2023, qin2023tool, gou2024critic}. These methods have shown promising results in significantly reducing hallucinated content and enhancing the accuracy, reliability, and faithfulness of model outputs. However, we caution that these approaches are not without limitations.

For example, the effectiveness of RAG-based methods heavily relies on the quality of their retrieval mechanisms \citep{Wu2024}. Moreover, accurately parsing and understanding the content of input queries poses a challenge, especially when queries are inherently ambiguous or irrelevant to the domain of focus~\citep{tonmoy2024comprehensive}. Additionally, the task of retrieving relevant and precise information from extensive corpora can be computationally demanding and resource-intensive, necessitating continuous updating and modification of knowledge databases to keep pace with the latest information~\citep{chen2023benchmarking,siriwardhana2023improving,ram2023context,cheng2024lift}. There may also be situations where the knowledge database might contain conflicting or contradictory information, making it unclear which pieces of relevant information to extract~\citep{wang2023survey,yu2023chain,gao2023retrieval}. For instance, when a legal case is overruled, or when there is a circuit split on an issue, the retrieval module must have some mechanism to distinguish outdated or jursidictionally-irrelevant sources from apposite and binding law.

Furthermore, methods for detecting hallucinations and evaluating their mitigation are themselves not foolproof. Evaluation datasets and metrics may not always accurately reflect real-world performance reliability~\citep{Ji2023,Zhang2023,lucas-etal-2023-fighting}. Biases could be embedded within the evaluation dataset, or the automated metric employed to quantify hallucination may lack comprehensiveness or task-specificity~\citep{kang2024comparing}. Therefore, it is far from clear whether these technical improvements will be able to truly solve the hallucination problem.

Finally, we also emphasize that the challenges presented by legal hallucinations are not only empirical, but also normative. Although data-rich and moneyed players certainly stand at an advantage when it comes to building hallucination-free legal LLMs for their own private use, it is not clear that even infinite resources can entirely solve the conceptual problems we diagnose. As we discuss in Section~\ref{sec:hallucination_trade_offs}, model fidelity to the training corpus, model fidelity to the user's prompt, and model fidelity to the facts of the world---i.e., the law---are normative commitments that stand in tension with each other, despite all being independently desirable technical properties of an LLM. Ultimately, since hallucinations of \textit{some} kind are generally inevitable at the token level \citep{Kalai2023, Xu2024}, developers of legal LLMs will need to make choices about which type(s) of hallucinations to minimize, and they should make these choices transparent to their downstream users. Only then can individual litigants decide for themselves whether the legal information they seek to obtain from LLMs is trustworthy or not.

To aid in future research in this area, we release a test dataset of our queries and answers on the HuggingFace platform, which scholars can use to continue to evaluate LLMs as they advance in legal sophistication.\footnote{\url{https://huggingface.co/datasets/reglab/legal_hallucinations}} In the meantime, more experienced legal practitioners may find some value in consulting LLMs for certain tasks, but even these users should remain vigilant in their use, taking care to verify the accuracy of their prompts and the quality of their chosen LLM's responses. Similarly, legal scholars and educators seeking to use LLMs as automated research assistants \citep{Livermore2024} or student aids \citep{Choi2024a} must be cautious to not inadvertently inject these LLMs' subtle knowledge biases into their own applications. Put differently, our findings underscore the importance of human-centered AI. Responsible integration of LLMs into legal tasks must \textit{augment} lawyers, researchers, and litigants and not, as Chief Justice Roberts has put it, risk ``dehumanizing the law'' \citep[5]{Roberts2023}.

\clearpage
\begin{table*}

\caption{\label{tab:CategoryOneTable}Hallucination rates across levels of the federal judiciary (low complexity tasks)}
\centering
\resizebox{\linewidth}{!}{
\begin{threeparttable}
\begin{tabular}[t]{llllllllllllll}
\toprule
\multicolumn{2}{c}{ } & \multicolumn{4}{c}{\makecell[c]{SCOTUS\\(1794-2015; n=5000)}} & \multicolumn{4}{c}{\makecell[c]{USCOA\\(1895-2019; n=5000)}} & \multicolumn{4}{c}{\makecell[c]{USDC\\(1932-2019; n=5000)}} \\
\cmidrule(l{3pt}r{3pt}){3-6} \cmidrule(l{3pt}r{3pt}){7-10} \cmidrule(l{3pt}r{3pt}){11-14}
Task & Prompt & GPT 4 & GPT 3.5 & PaLM 2 & Llama 2 & GPT 4 & GPT 3.5 & PaLM 2 & Llama 2 & GPT 4 & GPT 3.5 & PaLM 2 & Llama 2\\
\midrule
 & Zero-shot & \makecell[c]{0.204\\(0.006)} & \makecell[c]{0.004\\(0.001)} & \makecell[c]{0.054\\(0.003)} & \makecell[c]{0.303\\(0.006)} & \makecell[c]{0.237\\(0.006)} & \makecell[c]{0.003\\(0.001)} & \makecell[c]{0.025\\(0.002)} & \makecell[c]{0.157\\(0.005)} & \makecell[c]{0.174\\(0.005)} & \makecell[c]{0.001\\(0.001)} & \makecell[c]{0.016\\(0.002)} & \makecell[c]{0.240\\(0.006)}\\
\cmidrule{2-14}
\multirow{-2}{*}[1em]{\raggedright\arraybackslash Existence} & Few-shot & \makecell[c]{0.181\\(0.005)} & \makecell[c]{0.029\\(0.002)} & \makecell[c]{0.029\\(0.002)} & \makecell[c]{1.000\\(0.000)} & \makecell[c]{0.129\\(0.005)} & \makecell[c]{0.018\\(0.002)} & \makecell[c]{0.005\\(0.001)} & \makecell[c]{1.000\\(0.000)} & \makecell[c]{0.048\\(0.003)} & \makecell[c]{0.004\\(0.001)} & \makecell[c]{0.006\\(0.001)} & \makecell[c]{1.000\\(0.000)}\\
\cmidrule{1-14}
 & Zero-shot & \makecell[c]{0.000\\(0.000)} & \makecell[c]{0.000\\(0.000)} & \makecell[c]{0.000\\(0.000)} & \makecell[c]{0.003\\(0.001)} & \makecell[c]{0.490\\(0.007)} & \makecell[c]{0.645\\(0.007)} & \makecell[c]{0.703\\(0.006)} & \makecell[c]{0.700\\(0.006)} & \makecell[c]{0.829\\(0.005)} & \makecell[c]{0.815\\(0.005)} & \makecell[c]{0.839\\(0.005)} & \makecell[c]{0.815\\(0.005)}\\
\cmidrule{2-14}
\multirow{-2}{*}[1em]{\raggedright\arraybackslash Court} & Few-shot & \makecell[c]{0.000\\(0.000)} & \makecell[c]{0.000\\(0.000)} & \makecell[c]{0.000\\(0.000)} & \makecell[c]{0.005\\(0.001)} & \makecell[c]{0.491\\(0.007)} & \makecell[c]{0.641\\(0.007)} & \makecell[c]{0.703\\(0.006)} & \makecell[c]{0.679\\(0.007)} & \makecell[c]{0.831\\(0.005)} & \makecell[c]{0.870\\(0.005)} & \makecell[c]{0.842\\(0.005)} & \makecell[c]{0.870\\(0.005)}\\
\cmidrule{1-14}
 & Zero-shot & \makecell[c]{0.621\\(0.007)} & \makecell[c]{0.684\\(0.007)} & \makecell[c]{0.906\\(0.004)} & \makecell[c]{0.941\\(0.003)} & \makecell[c]{0.727\\(0.006)} & \makecell[c]{0.754\\(0.006)} & \makecell[c]{1.000\\(0.000)} & \makecell[c]{0.999\\(0.000)} & \makecell[c]{0.610\\(0.007)} & \makecell[c]{0.702\\(0.006)} & \makecell[c]{1.000\\(0.000)} & \makecell[c]{1.000\\(0.000)}\\
\cmidrule{2-14}
\multirow{-2}{*}[1em]{\raggedright\arraybackslash Citation} & Few-shot & \makecell[c]{0.689\\(0.007)} & \makecell[c]{0.641\\(0.007)} & \makecell[c]{0.833\\(0.005)} & \makecell[c]{0.953\\(0.003)} & \makecell[c]{0.468\\(0.007)} & \makecell[c]{0.920\\(0.004)} & \makecell[c]{0.999\\(0.001)} & \makecell[c]{1.000\\(0.000)} & \makecell[c]{0.270\\(0.006)} & \makecell[c]{0.942\\(0.003)} & \makecell[c]{0.999\\(0.000)} & \makecell[c]{1.000\\(0.000)}\\
\cmidrule{1-14}
 & Zero-shot & \makecell[c]{0.799\\(0.006)} & \makecell[c]{0.796\\(0.006)} & \makecell[c]{0.816\\(0.005)} & \makecell[c]{0.884\\(0.005)} & \makecell[c]{0.954\\(0.003)} & \makecell[c]{0.965\\(0.003)} & \makecell[c]{0.988\\(0.002)} & \makecell[c]{0.991\\(0.001)} & \makecell[c]{0.922\\(0.004)} & \makecell[c]{0.911\\(0.004)} & \makecell[c]{0.988\\(0.002)} & \makecell[c]{0.987\\(0.002)}\\
\cmidrule{2-14}
\multirow{-2}{*}[1em]{\raggedright\arraybackslash Author} & Few-shot & \makecell[c]{0.799\\(0.006)} & \makecell[c]{0.830\\(0.005)} & \makecell[c]{0.859\\(0.005)} & \makecell[c]{0.881\\(0.005)} & \makecell[c]{0.962\\(0.003)} & \makecell[c]{0.967\\(0.003)} & \makecell[c]{0.988\\(0.002)} & \makecell[c]{0.993\\(0.001)} & \makecell[c]{0.921\\(0.004)} & \makecell[c]{0.941\\(0.003)} & \makecell[c]{0.984\\(0.002)} & \makecell[c]{0.987\\(0.002)}\\
\bottomrule
\end{tabular}
\begin{tablenotes}[para]
\item \\ \textit{Note:} Table reports estimated hallucination rates. Standard errors are shown in parentheses.
\end{tablenotes}
\end{threeparttable}}
\end{table*} 

\begin{table*}

\caption{\label{tab:CategoryTwoTable}Hallucination rates across levels of the federal judiciary (moderate complexity tasks)}
\centering
\resizebox{\linewidth}{!}{
\begin{threeparttable}
\begin{tabular}[t]{llllllllllllll}
\toprule
\multicolumn{2}{c}{ } & \multicolumn{4}{c}{\makecell[c]{SCOTUS\\(1794-2015; n=5000)}} & \multicolumn{4}{c}{\makecell[c]{USCOA\\(1895-2019; n=5000)}} & \multicolumn{4}{c}{\makecell[c]{USDC\\(1932-2019; n=5000)}} \\
\cmidrule(l{3pt}r{3pt}){3-6} \cmidrule(l{3pt}r{3pt}){7-10} \cmidrule(l{3pt}r{3pt}){11-14}
Task & Prompt & GPT 4 & GPT 3.5 & PaLM 2 & Llama 2 & GPT 4 & GPT 3.5 & PaLM 2 & Llama 2 & GPT 4 & GPT 3.5 & PaLM 2 & Llama 2\\
\midrule
 & Zero-shot & \makecell[c]{0.399\\(0.007)} & \makecell[c]{0.499\\(0.001)} & \makecell[c]{0.500\\(0.000)} & \makecell[c]{0.536\\(0.007)} & \makecell[c]{0.494\\(0.005)} & \makecell[c]{0.500\\(0.000)} & \makecell[c]{0.500\\(0.000)} & \makecell[c]{0.493\\(0.008)} & - & - & - & -\\
\cmidrule{2-14}
\multirow{-2}{*}[1em]{\raggedright\arraybackslash Disposition} & Few-shot & \makecell[c]{0.452\\(0.007)} & \makecell[c]{0.496\\(0.002)} & \makecell[c]{0.501\\(0.002)} & \makecell[c]{0.502\\(0.004)} & \makecell[c]{0.498\\(0.006)} & \makecell[c]{0.489\\(0.007)} & \makecell[c]{0.501\\(0.001)} & \makecell[c]{0.501\\(0.002)} & - & - & - & -\\
\cmidrule{1-14}
 & Zero-shot & \makecell[c]{0.312\\(0.007)} & \makecell[c]{0.229\\(0.006)} & \makecell[c]{0.993\\(0.001)} & \makecell[c]{0.920\\(0.004)} & \makecell[c]{0.001\\(0.000)} & \makecell[c]{0.000\\(0.000)} & \makecell[c]{0.999\\(0.000)} & \makecell[c]{0.993\\(0.001)} & \makecell[c]{0.001\\(0.000)} & \makecell[c]{0.000\\(0.000)} & \makecell[c]{0.994\\(0.001)} & \makecell[c]{0.964\\(0.003)}\\
\cmidrule{2-14}
\multirow{-2}{*}[1em]{\raggedright\arraybackslash Quotation} & Few-shot & \makecell[c]{0.854\\(0.005)} & \makecell[c]{1.000\\(0.000)} & \makecell[c]{0.993\\(0.001)} & \makecell[c]{0.992\\(0.001)} & \makecell[c]{0.637\\(0.007)} & \makecell[c]{1.000\\(0.000)} & \makecell[c]{0.997\\(0.001)} & \makecell[c]{1.000\\(0.000)} & \makecell[c]{0.743\\(0.006)} & \makecell[c]{1.000\\(0.000)} & \makecell[c]{1.000\\(0.000)} & \makecell[c]{1.000\\(0.000)}\\
\cmidrule{1-14}
 & Zero-shot & \makecell[c]{0.922\\(0.004)} & \makecell[c]{0.937\\(0.003)} & \makecell[c]{0.985\\(0.002)} & \makecell[c]{0.991\\(0.001)} & \makecell[c]{0.934\\(0.004)} & \makecell[c]{0.978\\(0.002)} & \makecell[c]{0.997\\(0.001)} & \makecell[c]{0.999\\(0.001)} & \makecell[c]{0.956\\(0.003)} & \makecell[c]{0.870\\(0.005)} & \makecell[c]{0.995\\(0.001)} & \makecell[c]{0.999\\(0.000)}\\
\cmidrule{2-14}
\multirow{-2}{*}[1em]{\raggedright\arraybackslash Authority} & Few-shot & \makecell[c]{0.828\\(0.005)} & \makecell[c]{0.916\\(0.004)} & \makecell[c]{0.953\\(0.003)} & \makecell[c]{0.993\\(0.001)} & \makecell[c]{0.958\\(0.003)} & \makecell[c]{0.976\\(0.002)} & \makecell[c]{0.993\\(0.001)} & \makecell[c]{0.999\\(0.000)} & \makecell[c]{0.928\\(0.004)} & \makecell[c]{0.956\\(0.003)} & \makecell[c]{0.994\\(0.001)} & \makecell[c]{0.997\\(0.001)}\\
\cmidrule{1-14}
 & Zero-shot & \makecell[c]{0.810\\(0.025)} & \makecell[c]{0.919\\(0.017)} & \makecell[c]{0.858\\(0.022)} & \makecell[c]{0.972\\(0.011)} & - & - & - & - & - & - & - & -\\
\cmidrule{2-14}
\multirow{-2}{*}[1em]{\raggedright\arraybackslash Overruling year\textsuperscript{a}} & Few-shot & \makecell[c]{0.725\\(0.028)} & \makecell[c]{0.976\\(0.010)} & \makecell[c]{0.870\\(0.021)} & \makecell[c]{0.984\\(0.008)} & - & - & - & - & - & - & - & -\\
\bottomrule
\end{tabular}
\begin{tablenotes}[para]
\item[a] 1810-2022 (n=279)
\item \\ \textit{Note:} Table reports estimated hallucination rates. Standard errors are shown in parentheses.
\end{tablenotes}
\end{threeparttable}}
\end{table*} 

\begin{table*}

\caption{\label{tab:CategoryThreeTable}Hallucination rates across levels of the federal judiciary (high complexity tasks)}
\centering
\resizebox{\linewidth}{!}{
\begin{threeparttable}
\begin{tabular}[t]{llllllllllllll}
\toprule
\multicolumn{2}{c}{ } & \multicolumn{4}{c}{\makecell[c]{SCOTUS\\(1794-2015; n=100)}} & \multicolumn{4}{c}{\makecell[c]{USCOA\\(1895-2019; n=100)}} & \multicolumn{4}{c}{\makecell[c]{USDC\\(1932-2019; n=100)}} \\
\cmidrule(l{3pt}r{3pt}){3-6} \cmidrule(l{3pt}r{3pt}){7-10} \cmidrule(l{3pt}r{3pt}){11-14}
Task & Prompt & GPT 4 & GPT 3.5 & PaLM 2 & Llama 2 & GPT 4 & GPT 3.5 & PaLM 2 & Llama 2 & GPT 4 & GPT 3.5 & PaLM 2 & Llama 2\\
\midrule
 & Zero-shot & \makecell[c]{0.461\\(0.005)} & \makecell[c]{0.500\\(0.000)} & \makecell[c]{0.466\\(0.005)} & \makecell[c]{0.500\\(0.000)} & - & - & - & - & - & - & - & -\\
\cmidrule{2-14}
\multirow{-2}{*}[1em]{\raggedright\arraybackslash Doctrinal agreement\textsuperscript{a}} & Few-shot & \makecell[c]{0.449\\(0.007)} & \makecell[c]{0.458\\(0.004)} & \makecell[c]{0.453\\(0.006)} & \makecell[c]{0.500\\(0.000)} & - & - & - & - & - & - & - & -\\
\cmidrule{1-14}
Factual background & Zero-shot & \makecell[c]{0.460\\(0.050)} & \makecell[c]{0.700\\(0.046)} & \makecell[c]{0.950\\(0.022)} & \makecell[c]{0.830\\(0.038)} & \makecell[c]{0.750\\(0.043)} & \makecell[c]{0.810\\(0.039)} & \makecell[c]{0.880\\(0.032)} & \makecell[c]{0.880\\(0.032)} & \makecell[c]{0.710\\(0.045)} & \makecell[c]{0.670\\(0.047)} & \makecell[c]{0.880\\(0.032)} & \makecell[c]{0.880\\(0.032)}\\
\cmidrule{1-14}
Procedural posture & Zero-shot & \makecell[c]{0.470\\(0.050)} & \makecell[c]{0.650\\(0.048)} & \makecell[c]{0.740\\(0.044)} & \makecell[c]{0.850\\(0.036)} & \makecell[c]{0.660\\(0.047)} & \makecell[c]{0.600\\(0.049)} & \makecell[c]{0.680\\(0.047)} & \makecell[c]{0.830\\(0.038)} & \makecell[c]{0.730\\(0.044)} & \makecell[c]{0.550\\(0.050)} & \makecell[c]{0.810\\(0.039)} & \makecell[c]{0.890\\(0.031)}\\
\cmidrule{1-14}
Subsequent history & Zero-shot & \makecell[c]{0.080\\(0.027)} & \makecell[c]{0.220\\(0.041)} & \makecell[c]{0.810\\(0.039)} & \makecell[c]{0.770\\(0.042)} & \makecell[c]{0.450\\(0.050)} & \makecell[c]{0.360\\(0.048)} & \makecell[c]{0.750\\(0.043)} & \makecell[c]{0.620\\(0.049)} & \makecell[c]{0.480\\(0.050)} & \makecell[c]{0.300\\(0.046)} & \makecell[c]{0.730\\(0.044)} & \makecell[c]{0.670\\(0.047)}\\
\cmidrule{1-14}
Core legal question & Zero-shot & \makecell[c]{0.570\\(0.050)} & \makecell[c]{0.860\\(0.035)} & \makecell[c]{0.870\\(0.034)} & \makecell[c]{0.920\\(0.027)} & \makecell[c]{0.810\\(0.039)} & \makecell[c]{0.880\\(0.032)} & \makecell[c]{0.950\\(0.022)} & \makecell[c]{0.960\\(0.020)} & \makecell[c]{0.760\\(0.043)} & \makecell[c]{0.720\\(0.045)} & \makecell[c]{0.920\\(0.027)} & \makecell[c]{0.890\\(0.031)}\\
\cmidrule{1-14}
Central holding & Zero-shot & \makecell[c]{0.600\\(0.049)} & \makecell[c]{0.730\\(0.044)} & \makecell[c]{0.780\\(0.041)} & \makecell[c]{0.920\\(0.027)} & \makecell[c]{0.830\\(0.038)} & \makecell[c]{0.870\\(0.034)} & \makecell[c]{0.950\\(0.022)} & \makecell[c]{0.950\\(0.022)} & \makecell[c]{0.770\\(0.042)} & \makecell[c]{0.730\\(0.044)} & \makecell[c]{0.930\\(0.026)} & \makecell[c]{0.840\\(0.037)}\\
\bottomrule
\end{tabular}
\begin{tablenotes}[para]
\item[a] 1796-2005 (n=5000)
\item \\ \textit{Note:} Table reports estimated hallucination rates. For all tasks except doctrinal agreement, this rate is only a lower bound on the true population rate. Standard errors are shown in parentheses.
\end{tablenotes}
\end{threeparttable}}
\end{table*} 

\begin{table*}

\caption{\label{tab:ContraryToFactTable}Hallucination rates across levels of the federal judiciary (contra-factual tasks)}
\centering
\resizebox{\linewidth}{!}{
\begin{threeparttable}
\begin{tabular}[t]{llllllllll}
\toprule
\multicolumn{2}{c}{ } & \multicolumn{4}{c}{\makecell[c]{SCOTUS\\(1794-2015; n=1000)}} & \multicolumn{4}{c}{\makecell[c]{USCOA\\(1895-2019; n=1000)}} \\
\cmidrule(l{3pt}r{3pt}){3-6} \cmidrule(l{3pt}r{3pt}){7-10}
Task & Prompt & GPT 4 & GPT 3.5 & PaLM 2 & Llama 2 & GPT 4 & GPT 3.5 & PaLM 2 & Llama 2\\
\midrule
False dissent premise & Zero-shot & \makecell[c]{0.691\\(0.015)} & \makecell[c]{0.338\\(0.015)} & \makecell[c]{0.990\\(0.003)} & \makecell[c]{0.000\\(0.000)} & \makecell[c]{0.842\\(0.012)} & \makecell[c]{0.408\\(0.016)} & \makecell[c]{0.983\\(0.004)} & \makecell[c]{0.021\\(0.005)}\\
False overruling premise & Zero-shot & \makecell[c]{0.531\\(0.016)} & \makecell[c]{0.821\\(0.012)} & \makecell[c]{1.000\\(0.000)} & \makecell[c]{0.027\\(0.005)} & - & - & - & -\\
\bottomrule
\end{tabular}
\begin{tablenotes}[para]
\item \\ \textit{Note:} Table reports estimated hallucination rates. Standard errors are shown in parentheses.
\end{tablenotes}
\end{threeparttable}}
\end{table*} 

\begin{table*}

\caption{\label{tab:ECETable}Expected calibration error (ECE) across levels of the federal judiciary}
\centering
\resizebox{\linewidth}{!}{
\begin{threeparttable}
\begin{tabular}[t]{llllllllllllll}
\toprule
\multicolumn{2}{c}{ } & \multicolumn{4}{c}{\makecell[c]{SCOTUS\\(1794-2015; n=100)}} & \multicolumn{4}{c}{\makecell[c]{USCOA\\(1895-2019; n=100)}} & \multicolumn{4}{c}{\makecell[c]{USDC\\(1932-2019; n=100)}} \\
\cmidrule(l{3pt}r{3pt}){3-6} \cmidrule(l{3pt}r{3pt}){7-10} \cmidrule(l{3pt}r{3pt}){11-14}
Task & Prompt & GPT 4 & GPT 3.5 & PaLM 2 & Llama 2 & GPT 4 & GPT 3.5 & PaLM 2 & Llama 2 & GPT 4 & GPT 3.5 & PaLM 2 & Llama 2\\
\midrule
 & Zero-shot & \makecell[c]{0.126\\(0.005)} & \makecell[c]{0.008\\(0.001)} & \makecell[c]{0.119\\(0.004)} & \makecell[c]{0.262\\(0.006)} & \makecell[c]{0.124\\(0.005)} & \makecell[c]{0.007\\(0.001)} & \makecell[c]{0.182\\(0.003)} & \makecell[c]{0.117\\(0.005)} & \makecell[c]{0.074\\(0.004)} & \makecell[c]{0.004\\(0.000)} & \makecell[c]{0.183\\(0.003)} & \makecell[c]{0.178\\(0.006)}\\
\cmidrule{2-14}
\multirow{-2}{*}[1em]{\raggedright\arraybackslash Existence} & Few-shot & \makecell[c]{0.149\\(0.005)} & \makecell[c]{0.038\\(0.002)} & \makecell[c]{0.041\\(0.002)} & \makecell[c]{0.998\\(0.000)} & \makecell[c]{0.073\\(0.005)} & \makecell[c]{0.064\\(0.002)} & \makecell[c]{0.014\\(0.001)} & \makecell[c]{0.998\\(0.000)} & \makecell[c]{0.024\\(0.003)} & \makecell[c]{0.053\\(0.002)} & \makecell[c]{0.038\\(0.001)} & \makecell[c]{0.996\\(0.000)}\\
\cmidrule{1-14}
 & Zero-shot & \makecell[c]{0.000\\(0.000)} & \makecell[c]{0.000\\(0.000)} & \makecell[c]{0.000\\(0.000)} & \makecell[c]{0.003\\(0.000)} & \makecell[c]{0.362\\(0.006)} & \makecell[c]{0.322\\(0.006)} & \makecell[c]{0.184\\(0.006)} & \makecell[c]{0.387\\(0.006)} & \makecell[c]{0.133\\(0.005)} & \makecell[c]{0.163\\(0.005)} & \makecell[c]{0.132\\(0.005)} & \makecell[c]{0.163\\(0.006)}\\
\cmidrule{2-14}
\multirow{-2}{*}[1em]{\raggedright\arraybackslash Court} & Few-shot & \makecell[c]{0.000\\(0.000)} & \makecell[c]{0.001\\(0.000)} & \makecell[c]{0.000\\(0.000)} & \makecell[c]{0.009\\(0.001)} & \makecell[c]{0.345\\(0.006)} & \makecell[c]{0.355\\(0.006)} & \makecell[c]{0.161\\(0.006)} & \makecell[c]{0.452\\(0.008)} & \makecell[c]{0.145\\(0.005)} & \makecell[c]{0.099\\(0.004)} & \makecell[c]{0.094\\(0.005)} & \makecell[c]{0.099\\(0.005)}\\
\cmidrule{1-14}
 & Zero-shot & \makecell[c]{0.113\\(0.005)} & \makecell[c]{0.069\\(0.004)} & \makecell[c]{0.026\\(0.003)} & \makecell[c]{0.068\\(0.003)} & \makecell[c]{0.143\\(0.004)} & \makecell[c]{0.073\\(0.005)} & \makecell[c]{0.004\\(0.001)} & \makecell[c]{0.036\\(0.001)} & \makecell[c]{0.104\\(0.003)} & \makecell[c]{0.043\\(0.003)} & \makecell[c]{0.002\\(0.000)} & \makecell[c]{0.022\\(0.001)}\\
\cmidrule{2-14}
\multirow{-2}{*}[1em]{\raggedright\arraybackslash Citation} & Few-shot & \makecell[c]{0.191\\(0.005)} & \makecell[c]{0.029\\(0.003)} & \makecell[c]{0.059\\(0.004)} & \makecell[c]{0.063\\(0.003)} & \makecell[c]{0.138\\(0.005)} & \makecell[c]{0.036\\(0.003)} & \makecell[c]{0.010\\(0.001)} & \makecell[c]{0.032\\(0.001)} & \makecell[c]{0.099\\(0.004)} & \makecell[c]{0.018\\(0.002)} & \makecell[c]{0.001\\(0.000)} & \makecell[c]{0.051\\(0.002)}\\
\cmidrule{1-14}
 & Zero-shot & \makecell[c]{0.443\\(0.006)} & \makecell[c]{0.208\\(0.006)} & \makecell[c]{0.130\\(0.005)} & \makecell[c]{0.419\\(0.006)} & \makecell[c]{0.349\\(0.005)} & \makecell[c]{0.094\\(0.003)} & \makecell[c]{0.131\\(0.002)} & \makecell[c]{0.583\\(0.004)} & \makecell[c]{0.302\\(0.005)} & \makecell[c]{0.228\\(0.005)} & \makecell[c]{0.063\\(0.002)} & \makecell[c]{0.305\\(0.004)}\\
\cmidrule{2-14}
\multirow{-2}{*}[1em]{\raggedright\arraybackslash Author} & Few-shot & \makecell[c]{0.454\\(0.006)} & \makecell[c]{0.347\\(0.006)} & \makecell[c]{0.142\\(0.005)} & \makecell[c]{0.454\\(0.005)} & \makecell[c]{0.357\\(0.004)} & \makecell[c]{0.142\\(0.003)} & \makecell[c]{0.118\\(0.003)} & \makecell[c]{0.656\\(0.004)} & \makecell[c]{0.320\\(0.004)} & \makecell[c]{0.096\\(0.003)} & \makecell[c]{0.045\\(0.002)} & \makecell[c]{0.481\\(0.005)}\\
\cmidrule{1-14}
 & Zero-shot & \makecell[c]{0.203\\(0.007)} & \makecell[c]{0.431\\(0.007)} & \makecell[c]{0.291\\(0.008)} & \makecell[c]{0.199\\(0.007)} & \makecell[c]{0.170\\(0.007)} & \makecell[c]{0.557\\(0.006)} & \makecell[c]{0.382\\(0.006)} & \makecell[c]{0.148\\(0.006)} & - & - & - & -\\
\cmidrule{2-14}
\multirow{-2}{*}[1em]{\raggedright\arraybackslash Disposition} & Few-shot & \makecell[c]{0.239\\(0.006)} & \makecell[c]{0.314\\(0.008)} & \makecell[c]{0.165\\(0.007)} & \makecell[c]{0.283\\(0.008)} & \makecell[c]{0.165\\(0.007)} & \makecell[c]{0.205\\(0.007)} & \makecell[c]{0.227\\(0.007)} & \makecell[c]{0.439\\(0.007)} & - & - & - & -\\
\cmidrule{1-14}
 & Zero-shot & \makecell[c]{0.308\\(0.025)} & \makecell[c]{0.246\\(0.022)} & \makecell[c]{0.116\\(0.019)} & \makecell[c]{0.510\\(0.019)} & - & - & - & - & - & - & - & -\\
\cmidrule{2-14}
\multirow{-2}{*}[1em]{\raggedright\arraybackslash Overruling year\textsuperscript{a}} & Few-shot & \makecell[c]{0.377\\(0.025)} & \makecell[c]{0.680\\(0.022)} & \makecell[c]{0.154\\(0.019)} & \makecell[c]{0.754\\(0.018)} & - & - & - & - & - & - & - & -\\
\cmidrule{1-14}
 & Zero-shot & \makecell[c]{0.369\\(0.006)} & \makecell[c]{0.527\\(0.007)} & \makecell[c]{0.165\\(0.006)} & \makecell[c]{0.564\\(0.006)} & - & - & - & - & - & - & - & -\\
\cmidrule{2-14}
\multirow{-2}{*}[1em]{\raggedright\arraybackslash Doctrinal agreement\textsuperscript{b}} & Few-shot & \makecell[c]{0.319\\(0.007)} & \makecell[c]{0.409\\(0.008)} & \makecell[c]{0.152\\(0.007)} & \makecell[c]{0.548\\(0.007)} & - & - & - & - & - & - & - & -\\
\bottomrule
\end{tabular}
\begin{tablenotes}[para]
\item[a] 1810-2022 (n=279)
\item[b] 1796-2005 (n=5000)
\item \\ \textit{Note:} Table reports expected calibration error between empirical hallucination rates and estimated conditional probabilities. Conditional probabilities are estimated by sampling 10 responses from the model at temperature 1 and assessing their agreement with the model"s greedy response. Bootstrapped standard errors are shown in parentheses.
\end{tablenotes}
\end{threeparttable}}
\end{table*}

\clearpage
\printbibliography[heading=bibintoc]
\end{refsection}

\appendix
\begin{refsection}

\clearpage
\begin{center}
    \Large Online Appendix
\end{center}
\clearpage

\setcounter{figure}{0}
\setcounter{table}{0}

\section{Data Sources} 
\label{appendix:data_sources}

In Section 4.1 in the main manuscript, we describe the data and sampling strategy that we use to construct our queries. For clarity, Appendix Table~\ref{tab:DataSourceTable} links these data to each task. Our exact procedure for sampling, merging, and aggregating across these datasets is available in our GitHub repository.

\begin{table*}[ht]

\caption{\label{tab:DataSourceTable}Data used to construct each task}
\centering
\begin{tabular}{lll}
\toprule
\textbf{Complexity} & \textbf{Task} & \textbf{Data source}\\
\midrule
 & Existence & CAP \citeyearpar{CaselawAccessProject2023}; SCDB \citep{Spaeth2022}\\
\cmidrule{2-3}
 & Court & CAP; SCDB\\
\cmidrule{2-3}
 & Citation & CAP; SCDB\\
\cmidrule{2-3}
\multirow{-4}{*}[0.75em]{\raggedright\arraybackslash Low} & Author & CAP; SCDB\\
\midrule
\cmidrule{1-3}
 & Disposition & CAP; SCDB; ACDB \citep{Songer2008, Kuersten2011}\\
\cmidrule{2-3}
 & Quotation & CAP; SCDB\\
\cmidrule{2-3}
 & Authority & CAP; SCDB\\
\cmidrule{2-3}
\multirow{-4}{*}[0.75em]{\raggedright\arraybackslash Moderate} & Overruling year & SCDB; Library of Congress \citeyearpar{Congress.gov2023}\\
\midrule
\cmidrule{1-3}
 & Doctrinal agreement & Shepard's \citep{Fowler2007, Black2013}\\
\cmidrule{2-3}
 & Factual background & CAP; SCDB\\
\cmidrule{2-3}
 & Procedural posture & CAP; SCDB\\
\cmidrule{2-3}
 & Subsequent history & CAP; SCDB\\
\cmidrule{2-3}
 & Core legal question & CAP; SCDB\\
\cmidrule{2-3}
\multirow{-6}{*}[1.15em]{\raggedright\arraybackslash High} & Central holding & CAP; SCDB\\
\bottomrule
\end{tabular}
\end{table*}

\section{Prompt Templates} 
\label{appendix:prompts}

The full zero-shot and few-shot prompt templates for all of our queries are shown in Appendix Figures~\ref{fig:first_prompt_template} to \ref{fig:last_prompt_template}. The few-shot examples presented are those used for the SCOTUS queries; appropriate cases from the other levels of the judiciary are used in the USCOA and USDC versions.

\section{Contradiction Detection Approach}
\label{appendix:contradiction_detection_approach}

\begin{table}[t]

\caption{\label{tab:IRRTable}Intercoder reliability}
\centering
\begin{tabular}[t]{ll}
\toprule
\textbf{Coders} & $\bm{\kappa}$\\
\midrule
Coder 1 vs. GPT 4 & 0.87\\
\cmidrule{1-2}
Coder 2 vs. GPT 4 & 0.77\\
\cmidrule{1-2}
Coder 1 vs. Coder 2 & 0.79\\
\bottomrule
\end{tabular}
\end{table}

In Section 4.3 in the main manuscript, we describe our strategy for assessing hallucinations for our reference-free tasks. Briefly, because we do not have access to ground-truth labels for these tasks, we exploit the stochastic behavior of LLMs at higher temperatures and check for contradictions in their responses to repeated queries. In Appendix Figure~\ref{fig:contradiction_elicitation_prompt}, we share the template for the contradiction elicitation query that we send to GPT 4 to perform this contradiction labeling at scale. We frame our contradiction check as an entailment task \citep{Wang2021} as we find that this prompting strategy is the most performant in our setting.

To confirm that GPT 4 is able to reliably detect the contradictions that we envision, we crosscheck GPT 4's conclusions against our own expert knowledge on a subsample of our data. Specifically, for 100 of our queries, two of us independently recoded the LLMs' responses for contradictions. We report Cohen's $\kappa$ coefficient \citep{Cohen1960} for our agreement with GPT-4 on these queries in Appendix Table~\ref{tab:IRRTable}. In general, a $\kappa$ value greater than 0.80 is considered ``almost perfect'' agreement, and one greater than 0.60 is considered ``substantial'' agreement \citep{Landis1977}. Our $\kappa$ values suggest that we are well-justified to use GPT 4 for contradiction detection; indeed, one of us agrees with GPT 4 \textit{more} than with the other human coder.

To give the reader a sense of GPT 4's reasoning abilities in this setting, in Appendix Figures~\ref{fig:gpt4_contradiction_label_positive} and \ref{fig:gpt4_contradiction_label_negative} we share some examples of its reasoning process and contradiction conclusions.

\section{Expected Calibration Error (ECE)}
\label{appendix:ece_estimation}

\subsection{Estimation Approach}

In Section 5.3 in the main manuscript, we estimate the expected calibration error (ECE) of our LLMs for each of our tasks. Studying model calibration directly is not possible in our setup because we do not always observe our LLMs' conditional probability distributions, which we require in order to determine whether they are in fact correlated with their response accuracies. Specifically, three of the models that we evaluate---OpenAI's ChatGPT 4, OpenAI's ChatGPT 3.5, and Google's PaLM 2---are closed-source and do not expose this information to the user. To overcome this hurdle, we instead \textit{estimate} the distributions by drawing $K$ samples from the model at temperature 1 and comparing those responses to its greedy response as follows:

\begin{equation}
\label{eq:est_conditional_probability_distribution}
    \widehat{\mbox{Pr}}[\text{response}|\cdot] = \frac{1}{K} \sum_{k \in K} \mathbbm{1}[\text{response} = f_{1}(\cdot)^{(k)}]
\end{equation}

Then, for each task, we calculate the ECE of the LLM. Conceptually, the ECE represents the average difference between the LLM's confidence in the accuracy of its responses and the empirical frequency of its correct, non-hallucinated responses. Formally, where $X = \widehat{\mbox{Pr}}[\text{response}|\cdot]$ (cf. Equation~\ref{eq:est_conditional_probability_distribution} above) and $Y = \mathbbm{1}[\text{response} = \text{response}^\prime]$ (cf. Equation 1 in the main manuscript), we estimate:

\begin{equation}
    \text{ECE} = \mathbbm{E}[| X - \mathbbm{E}[Y | X] |]
\end{equation}

Following \citet{Kumar2020}, we use a plug-in estimator that bins the data into 10 equally-sized slices and subtracts off an approximated bias term.

\subsection{Temperature Scaling}

As we report in Section 5.3 in the main manuscript, our results suggest that out of the box, the LLMs we evaluate are not well-calibrated on legal queries. However, a model that is uncalibrated under its unscaled probability distribution is not necessarily uncalibrated \textit{tout court}; the purpose of a LLM's temperature parameter, after all, is to allow an sophisticated user to adjust the distribution as needed. To explore whether such temperature scaling could indeed affect our results, here we perform \textit{ex post} Platt scaling \citep{Guo2017} on the raw distribution and check for improvements in the measured ECE.

\begin{figure}
    \centering
    \includegraphics[width=4in]{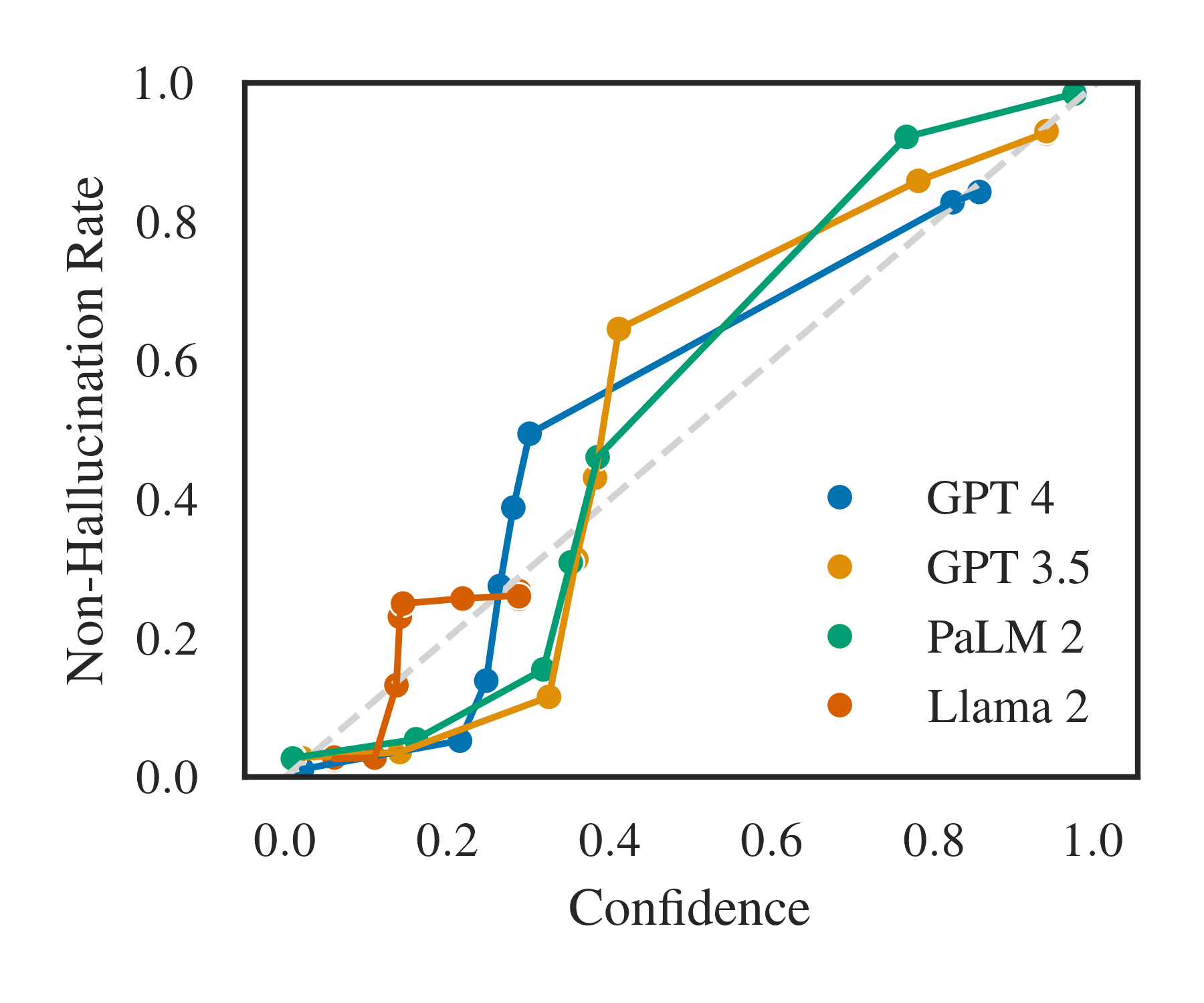}
    \caption{Rescaled calibration curves by LLM, all resource-aware tasks pooled.}
    \label{fig:ece_scaled}
\end{figure}

Appendix Figure~\ref{fig:ece_scaled} visualizes the rescaled ECE for each of our LLMs, Appendix Table~\ref{tab:ECEScaledTable} reports the task-level results, and Appendix Table~\ref{tab:ECEChangeTable} summarizes the numerical gains. As expected, rescaling generally improves the ECE, especially for GPT 4 and Llama 2. However, the rescaling procedure is not perfect: GPT4, GPT 3.5, and PaLM 2 remain relatively uncalibrated in the $[0.2, 0.4]$ confidence interval. And although the pooled ECE of Llama 2 improves substantially, this is due to the entire distribution being compressed into the $[0.0, 0.3]$ interval---the rescaled Llama 2 model is simply not confident in any of its responses. Overall, these results confirm our conclusion in Section 5.3 that LLMs face calibration challenges on legal knowledge queries, though they do suggest that \textit{ex post} rescaling can help.

\begin{table*}

\caption{\label{tab:ECEScaledTable}Temperature-scaled expected calibration error (ECE) across levels of the federal judiciary}
\centering
\resizebox{\linewidth}{!}{
\begin{threeparttable}
\begin{tabular}[t]{llllllllllllll}
\toprule
\multicolumn{2}{c}{ } & \multicolumn{4}{c}{\makecell[c]{SCOTUS\\(1794-2015; n=100)}} & \multicolumn{4}{c}{\makecell[c]{USCOA\\(1895-2019; n=100)}} & \multicolumn{4}{c}{\makecell[c]{USDC\\(1932-2019; n=100)}} \\
\cmidrule(l{3pt}r{3pt}){3-6} \cmidrule(l{3pt}r{3pt}){7-10} \cmidrule(l{3pt}r{3pt}){11-14}
Task & Prompt & GPT 4 & GPT 3.5 & PaLM 2 & Llama 2 & GPT 4 & GPT 3.5 & PaLM 2 & Llama 2 & GPT 4 & GPT 3.5 & PaLM 2 & Llama 2\\
\midrule
 & Zero-shot & \makecell[c]{0.002\\(0.003)} & \makecell[c]{0.000\\(0.001)} & \makecell[c]{0.000\\(0.002)} & \makecell[c]{0.001\\(0.005)} & \makecell[c]{0.003\\(0.004)} & \makecell[c]{0.000\\(0.000)} & \makecell[c]{0.000\\(0.001)} & \makecell[c]{0.006\\(0.004)} & \makecell[c]{0.003\\(0.004)} & \makecell[c]{0.000\\(0.000)} & \makecell[c]{0.000\\(0.001)} & \makecell[c]{0.001\\(0.004)}\\
\cmidrule{2-14}
\multirow{-2}{*}[1em]{\raggedright\arraybackslash Existence} & Few-shot & \makecell[c]{0.001\\(0.003)} & \makecell[c]{0.017\\(0.002)} & \makecell[c]{0.004\\(0.002)} & \makecell[c]{0.000\\(0.000)} & \makecell[c]{0.023\\(0.004)} & \makecell[c]{0.000\\(0.001)} & \makecell[c]{0.000\\(0.001)} & \makecell[c]{0.000\\(0.000)} & \makecell[c]{0.013\\(0.002)} & \makecell[c]{0.000\\(0.001)} & \makecell[c]{0.000\\(0.001)} & \makecell[c]{0.996\\(0.000)}\\
\cmidrule{1-14}
 & Zero-shot & \makecell[c]{0.000\\(0.000)} & \makecell[c]{0.000\\(0.000)} & \makecell[c]{0.000\\(0.000)} & \makecell[c]{0.000\\(0.000)} & \makecell[c]{0.002\\(0.003)} & \makecell[c]{0.003\\(0.004)} & \makecell[c]{0.015\\(0.005)} & \makecell[c]{0.052\\(0.005)} & \makecell[c]{0.018\\(0.004)} & \makecell[c]{0.043\\(0.005)} & \makecell[c]{0.067\\(0.005)} & \makecell[c]{0.043\\(0.004)}\\
\cmidrule{2-14}
\multirow{-2}{*}[1em]{\raggedright\arraybackslash Court} & Few-shot & \makecell[c]{0.000\\(0.000)} & \makecell[c]{0.001\\(0.000)} & \makecell[c]{0.000\\(0.000)} & \makecell[c]{0.000\\(0.000)} & \makecell[c]{0.010\\(0.004)} & \makecell[c]{0.003\\(0.004)} & \makecell[c]{0.018\\(0.005)} & \makecell[c]{0.011\\(0.005)} & \makecell[c]{0.019\\(0.003)} & \makecell[c]{0.017\\(0.004)} & \makecell[c]{0.021\\(0.005)} & \makecell[c]{0.017\\(0.003)}\\
\cmidrule{1-14}
 & Zero-shot & \makecell[c]{0.064\\(0.004)} & \makecell[c]{0.073\\(0.004)} & \makecell[c]{0.016\\(0.003)} & \makecell[c]{0.026\\(0.003)} & \makecell[c]{0.000\\(0.002)} & \makecell[c]{0.007\\(0.003)} & \makecell[c]{0.000\\(0.000)} & \makecell[c]{0.000\\(0.000)} & \makecell[c]{0.001\\(0.001)} & \makecell[c]{0.000\\(0.001)} & \makecell[c]{0.000\\(0.000)} & \makecell[c]{0.022\\(0.001)}\\
\cmidrule{2-14}
\multirow{-2}{*}[1em]{\raggedright\arraybackslash Citation} & Few-shot & \makecell[c]{0.016\\(0.004)} & \makecell[c]{0.100\\(0.004)} & \makecell[c]{0.024\\(0.003)} & \makecell[c]{0.030\\(0.002)} & \makecell[c]{0.000\\(0.001)} & \makecell[c]{0.009\\(0.002)} & \makecell[c]{0.000\\(0.000)} & \makecell[c]{0.000\\(0.000)} & \makecell[c]{0.001\\(0.001)} & \makecell[c]{0.002\\(0.001)} & \makecell[c]{0.000\\(0.000)} & \makecell[c]{0.000\\(0.000)}\\
\cmidrule{1-14}
 & Zero-shot & \makecell[c]{0.003\\(0.003)} & \makecell[c]{0.028\\(0.005)} & \makecell[c]{0.011\\(0.004)} & \makecell[c]{0.000\\(0.003)} & \makecell[c]{0.001\\(0.002)} & \makecell[c]{0.000\\(0.001)} & \makecell[c]{0.000\\(0.001)} & \makecell[c]{0.000\\(0.001)} & \makecell[c]{0.000\\(0.002)} & \makecell[c]{0.000\\(0.002)} & \makecell[c]{0.000\\(0.001)} & \makecell[c]{0.000\\(0.001)}\\
\cmidrule{2-14}
\multirow{-2}{*}[1em]{\raggedright\arraybackslash Author} & Few-shot & \makecell[c]{0.003\\(0.003)} & \makecell[c]{0.003\\(0.003)} & \makecell[c]{0.002\\(0.003)} & \makecell[c]{0.001\\(0.002)} & \makecell[c]{0.000\\(0.002)} & \makecell[c]{0.000\\(0.001)} & \makecell[c]{0.000\\(0.001)} & \makecell[c]{0.000\\(0.001)} & \makecell[c]{0.002\\(0.002)} & \makecell[c]{0.005\\(0.002)} & \makecell[c]{0.001\\(0.001)} & \makecell[c]{0.000\\(0.001)}\\
\cmidrule{1-14}
 & Zero-shot & \makecell[c]{0.003\\(0.004)} & \makecell[c]{0.000\\(0.004)} & \makecell[c]{0.000\\(0.004)} & \makecell[c]{0.001\\(0.004)} & \makecell[c]{0.000\\(0.004)} & \makecell[c]{0.000\\(0.004)} & \makecell[c]{0.000\\(0.004)} & \makecell[c]{0.001\\(0.004)} & - & - & - & -\\
\cmidrule{2-14}
\multirow{-2}{*}[1em]{\raggedright\arraybackslash Disposition} & Few-shot & \makecell[c]{0.000\\(0.004)} & \makecell[c]{0.000\\(0.005)} & \makecell[c]{0.000\\(0.004)} & \makecell[c]{0.000\\(0.005)} & \makecell[c]{0.000\\(0.004)} & \makecell[c]{0.001\\(0.004)} & \makecell[c]{0.000\\(0.004)} & \makecell[c]{0.000\\(0.004)} & - & - & - & -\\
\cmidrule{1-14}
 & Zero-shot & \makecell[c]{0.006\\(0.012)} & \makecell[c]{0.005\\(0.011)} & \makecell[c]{0.017\\(0.012)} & \makecell[c]{0.000\\(0.006)} & - & - & - & - & - & - & - & -\\
\cmidrule{2-14}
\multirow{-2}{*}[1em]{\raggedright\arraybackslash Overruling year\textsuperscript{a}} & Few-shot & \makecell[c]{0.042\\(0.016)} & \makecell[c]{0.000\\(0.006)} & \makecell[c]{0.031\\(0.016)} & \makecell[c]{0.000\\(0.005)} & - & - & - & - & - & - & - & -\\
\cmidrule{1-14}
 & Zero-shot & \makecell[c]{0.002\\(0.004)} & \makecell[c]{0.000\\(0.004)} & \makecell[c]{0.001\\(0.004)} & \makecell[c]{0.000\\(0.004)} & - & - & - & - & - & - & - & -\\
\cmidrule{2-14}
\multirow{-2}{*}[1em]{\raggedright\arraybackslash Doctrinal agreement\textsuperscript{b}} & Few-shot & \makecell[c]{0.001\\(0.004)} & \makecell[c]{0.000\\(0.004)} & \makecell[c]{0.001\\(0.004)} & \makecell[c]{0.000\\(0.004)} & - & - & - & - & - & - & - & -\\
\bottomrule
\end{tabular}
\begin{tablenotes}[para]
\item[a] 1810-2022 (n=279)
\item[b] 1796-2005 (n=5000)
\item \\ \textit{Note:} Table reports temperature-scaled expected calibration error between empirical hallucination rates and estimated conditional probabilities. Conditional probabilities are estimated by sampling 10 responses from the model at temperature 1 and assessing their agreement with the model"s greedy response. Bootstrapped standard errors are shown in parentheses.
\end{tablenotes}
\end{threeparttable}}
\end{table*} 

\begin{table}

\caption{\label{tab:ECEChangeTable}Scaled and unscaled ECE}
\centering
\begin{tabular}[t]{lll}
\toprule
\textbf{LLM} & \textbf{ECE (unscaled)} & \textbf{ECE (scaled)}\\
\midrule
GPT 4 & 0.190 & 0.033\\
\cmidrule{1-3}
GPT 3.5 & 0.099 & 0.051\\
\cmidrule{1-3}
PaLM 2 & 0.057 & 0.049\\
\cmidrule{1-3}
Llama 2 & 0.421 & 0.035\\
\bottomrule
\end{tabular}
\end{table}

\section{Supplementary Analyses}

\subsection{Fake Case Existence Task}
\label{appendix:fake_case_existence}

\begin{table*}

\caption{\label{tab:FakeExistenceTable}Hallucination rates across levels of the federal judiciary (fake existence task)}
\centering
\resizebox{\linewidth}{!}{
\begin{threeparttable}
\begin{tabular}[t]{llllllllllllll}
\toprule
\multicolumn{2}{c}{ } & \multicolumn{4}{c}{\makecell[c]{SCOTUS\\(1794-2015; n=1000)}} & \multicolumn{4}{c}{\makecell[c]{USCOA\\(1895-2019; n=1000)}} & \multicolumn{4}{c}{\makecell[c]{USDC\\(1932-2019; n=5000)}} \\
\cmidrule(l{3pt}r{3pt}){3-6} \cmidrule(l{3pt}r{3pt}){7-10} \cmidrule(l{3pt}r{3pt}){11-14}
Task & Prompt & GPT 4 & GPT 3.5 & PaLM 2 & Llama 2 & GPT 4 & GPT 3.5 & PaLM 2 & Llama 2 & GPT 4 & GPT 3.5 & PaLM 2 & Llama 2\\
\midrule
False existence & Zero-shot & \makecell[c]{0.000\\(0.000)} & \makecell[c]{0.661\\(0.015)} & \makecell[c]{0.171\\(0.012)} & \makecell[c]{0.000\\(0.000)} & \makecell[c]{0.034\\(0.006)} & \makecell[c]{0.799\\(0.013)} & \makecell[c]{0.674\\(0.015)} & \makecell[c]{0.000\\(0.000)} & \makecell[c]{0.066\\(0.008)} & \makecell[c]{0.968\\(0.006)} & \makecell[c]{0.706\\(0.014)} & \makecell[c]{0.000\\(0.000)}\\
\bottomrule
\end{tabular}
\begin{tablenotes}[para]
\item \textit{Note:} Table reports estimated hallucination rates. Standard errors are shown in parentheses.
\end{tablenotes}
\end{threeparttable}}
\end{table*}

In Table 2 in the main manuscript we report results for the \textbf{Existence} task, where our LLMs are asked whether or not a given case exists. Performance on this task is generally strong; however, because all our prompted cases are in fact real, it is unclear whether these results are due to the LLMs' genuine knowledge of a case's existence or simply their tendency to always answer ``yes'' to this type of question. Accordingly, here we conduct a supplemental analysis where we repeat the existence query, but using fake cases instead of real cases. Each fake case citation is constructed with plausible party names and the reporter that is appropriate for the court at issue, e.g., \texttt{SolarFlare Technologies v. Armstrong, 656 F.3d 262} for a fake USCOA case.

Appendix Table~\ref{tab:FakeExistenceTable} reports the results of this \textbf{Fake case existence} experiment. We see that GPT 3.5 and PaLM 2 are both prone to simply asserting the existence of any case---real or fake---though PaLM 2 is more discriminating at the SCTOUS level. GPT 4 and Llama 2, on the other hand, appear immune from this behavior, but recall from Table 2 that this is just bias in the \textit{opposite} direction: they simply prefer to deny the existence of any case, real or fake. Altogether, these results belie the LLMs' seemingly satisfactory performance on the \textbf{Existence} task: even here, they may not possess any actual knowledge of the true details of many cases.

\subsection{Hallucination Rates at the USCOA Level (No Time Cutoff)}
\label{appendix:coa_map_full}

\begin{figure}
    \centering
\includegraphics[width=5in]{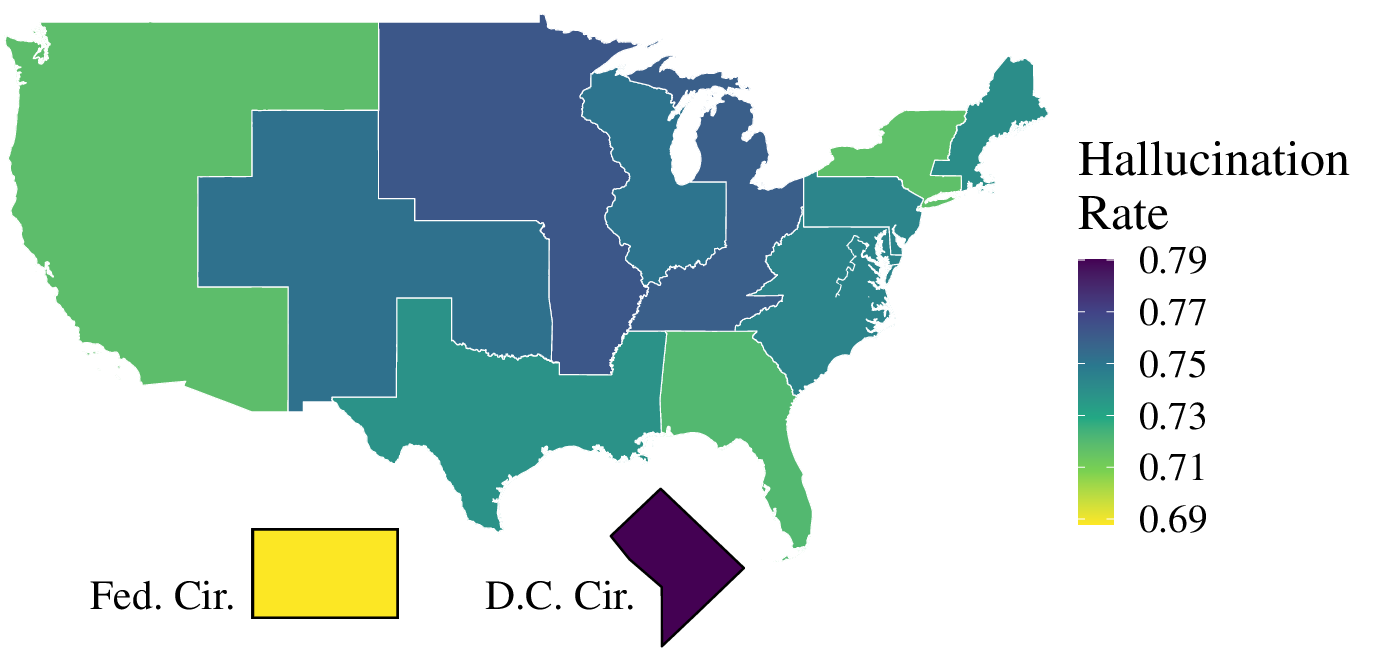}
    \caption{Relationship between USCOA jurisdiction and mean hallucination rate, all resource-aware USCOA tasks and models pooled. (No time cutoff.)}
    \label{fig:court_coa_geography}
\end{figure}

In Figure 4 in Section 5.1.3 in the main manuscript, we show the relationship between hallucinations and USCOA jurisdiction. However, in that figure, we exclude cases decided prior to 1982 in order to fairly compare rates across older and younger jurisdictions. Here, we share the results for that analysis with no cases excluded. As Appendix Figure~\ref{fig:court_coa_geography} suggests, the non-truncated results substantively mirror the truncated ones: the Ninth and Second Circuits continue to perform best. However, in this figure, the Federal and Eleventh Circuits have a somewhat stronger showing as well. We believe that this is best explained by these circuits' relative infancy: the Eleventh Circuit split from the Fifth Circuit in 1981, and the Federal Circuit was inaugurated in 1982, so they are composed of newer cases only.

\subsection{Hallucination Rates by State}
\label{appendix:usdc_map}

\begin{figure}
    \centering
\includegraphics[width=5in]{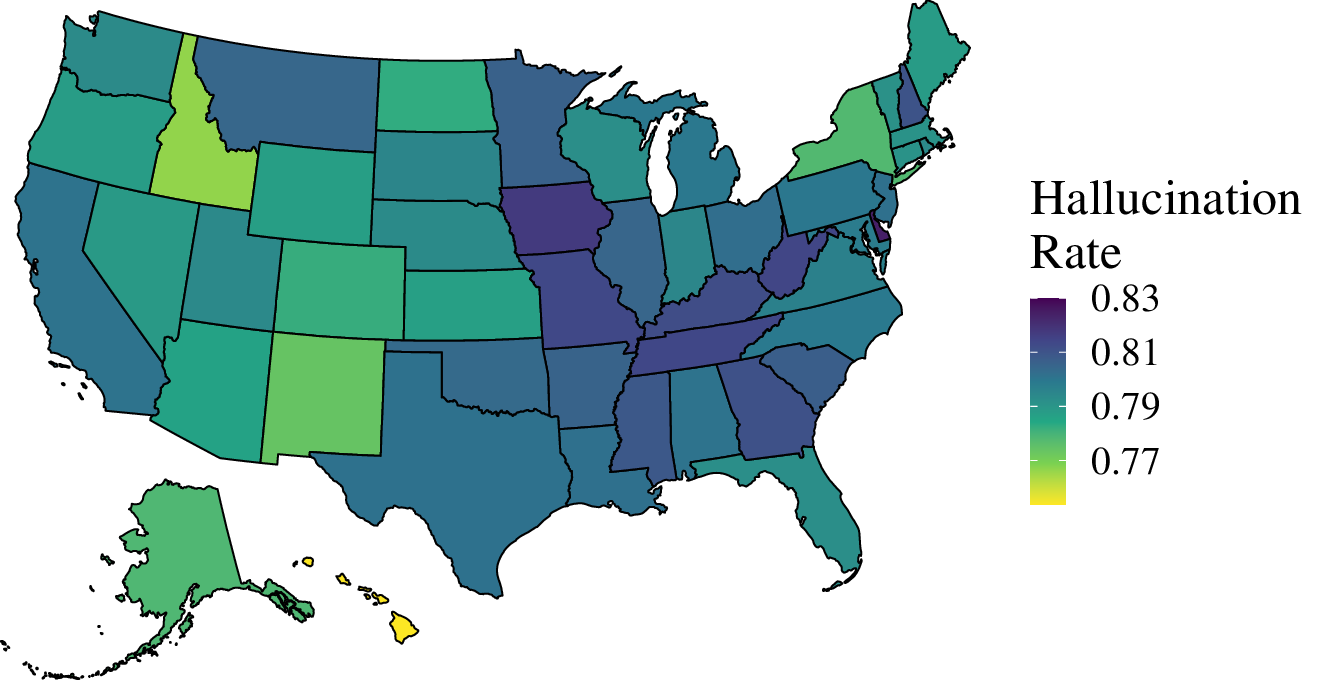}
    \caption{Relationship between aggregate state USDC jurisdiction and mean hallucination rate, all resource-aware USDC tasks and models pooled.}
    \label{fig:court_usdc_geography}
\end{figure}

In Section 5.1.3 in the main manuscript, we show within-USCOA sources of hallucination heterogeneity on a geographic map. Appendix Figure~\ref{fig:court_usdc_geography} depicts a similar analysis for our USDC tasks, aggregated to the state level. Confirming our results above, we observe a particularly low hallucination rate in the state of New York, which is comprised of the Northern District, Southern District, Western District, and Eastern District of New York. Interestingly, both Hawaii and Idaho (one district court each) also evince strong performance here.

\subsection{Zero-shot Prompting}

Most of the analysis in the main manuscript is focused on questions presented with three-shot prompting, or prompting that contains three examples of correct answers. We focus our analysis on this prompting approach for several reasons. First, few-shot prompting makes LLMs more likely to respond in the desired format, allowing for more reliable parsing of the LLM's response. Second, we are interested in uncovering the actual knowledge that is embedded in LLMs; a few-shot prompt better exposes this knowledge because it circumvents the general reticence that has been tuned into many commercial models. (For tasks where a non-hallucination requires the LLM to respond in the negative, like our contra-factual tasks, we do not use few-shot prompting because we are interested in the model's tuning and behavior rather than any information the LLM itself contains.) Finally, for many experienced users, a few-shot or conversational approach with many examples of the LLM responding correctly is the standard way to interact with these models for research and writing tasks, so this mode of evaluation most directly speaks to actual use cases.

In Tables 2, 3, and 4 in the main manuscript, we present the complete results for all of our queries in both the few-shot and zero-shot setups. Here, in Appendix Figure \ref{appendix:zeroshot}, we reproduce Figure 7 to show the summarized overall results for the zero-shot prompts alone. We see that the ordering of the models is preserved, but that the magnitude of each LLM's hallucination rate is reduced. These changes are largely driven by abstentions in the \textbf{Quotation} and the more complex tasks, which we count as non-hallucinations (see Appendix Section~\ref{appendix:abstension_rates} below). All of the trends that we discuss in the main manuscript (task complexity, court level, jurisdiction, case importance, and case year) continue to be present in our zero-shot results, but they are somewhat less pronounced in the zero-shot setting because of the higher abstention rates.

\begin{figure}
    \centering
    \includegraphics{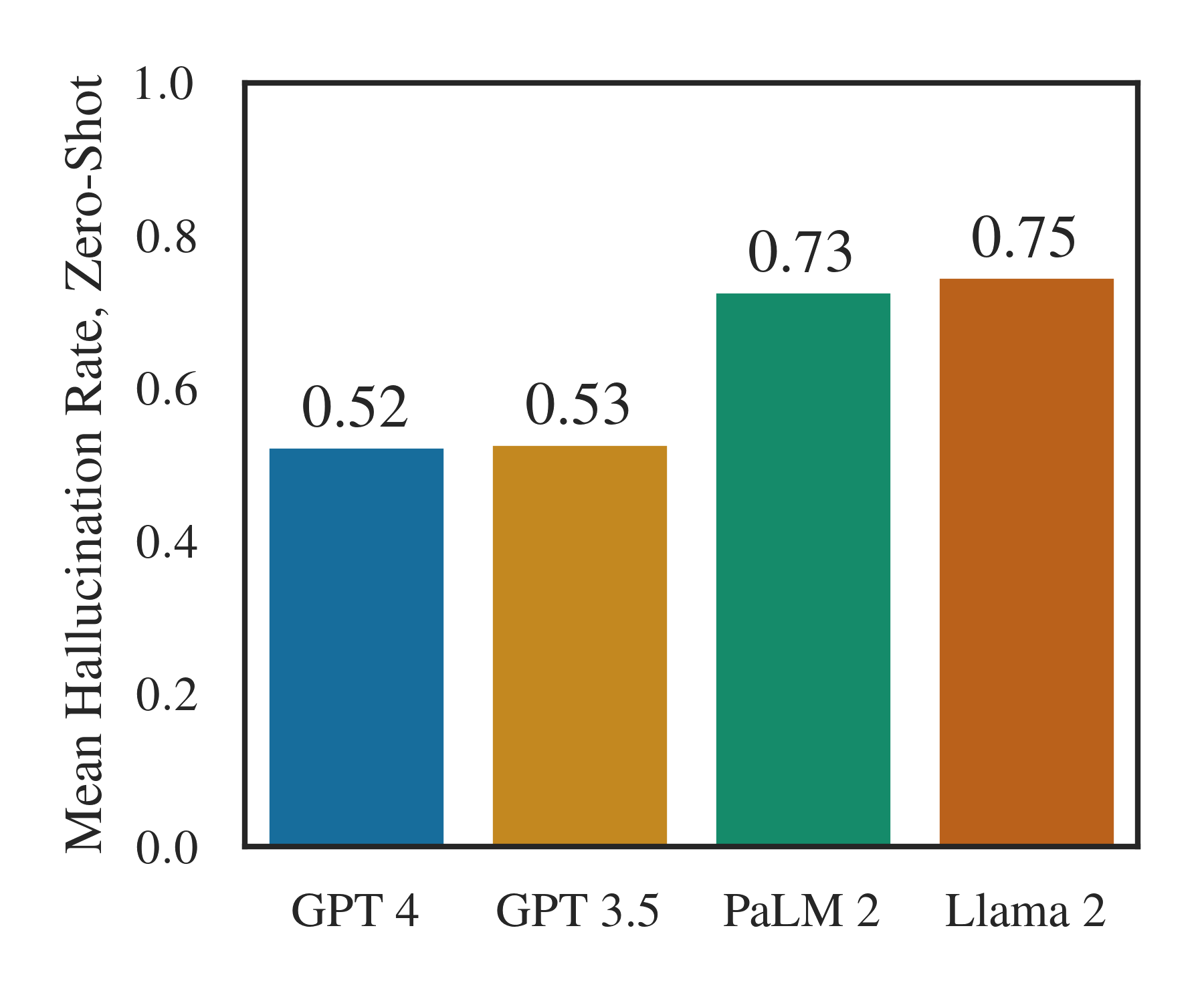}
    \caption{Hallucination rates by LLM, all reference-based tasks pooled, zero-shot prompts only.}
    \label{appendix:zeroshot}
\end{figure}

\section{Abstention Rates}
\label{appendix:abstension_rates}

Occasionally, our LLMs abstain from providing an answers to our queries. For example, they may plead ignorance or simply claim that they are unable to answer. When this occurs, we count these responses as non-hallucinations. Appendix Table~\ref{tab:NonResponseTable} reports the LLMs' abstention rates for each task, which are generally low. One exception is the zero-shot \textbf{Quotation} task, where many LLMs abstain.

\begin{table*}

\caption{\label{tab:NonResponseTable}Absention rates across levels of the federal judiciary (resource-aware tasks)}
\centering
\resizebox{\linewidth}{!}{
\begin{threeparttable}
\begin{tabular}[t]{llllllllllllll}
\toprule
\multicolumn{2}{c}{ } & \multicolumn{4}{c}{\makecell[c]{SCOTUS\\(1794-2015; n=100)}} & \multicolumn{4}{c}{\makecell[c]{USCOA\\(1895-2019; n=100)}} & \multicolumn{4}{c}{\makecell[c]{USDC\\(1932-2019; n=100)}} \\
\cmidrule(l{3pt}r{3pt}){3-6} \cmidrule(l{3pt}r{3pt}){7-10} \cmidrule(l{3pt}r{3pt}){11-14}
Task & Prompt & GPT 4 & GPT 3.5 & PaLM 2 & Llama 2 & GPT 4 & GPT 3.5 & PaLM 2 & Llama 2 & GPT 4 & GPT 3.5 & PaLM 2 & Llama 2\\
\midrule
 & Zero-shot & \makecell[c]{0.000\\(0.000)} & \makecell[c]{0.000\\(0.000)} & \makecell[c]{0.000\\(0.000)} & \makecell[c]{0.000\\(0.000)} & \makecell[c]{0.000\\(0.000)} & \makecell[c]{0.000\\(0.000)} & \makecell[c]{0.000\\(0.000)} & \makecell[c]{0.000\\(0.000)} & \makecell[c]{0.000\\(0.000)} & \makecell[c]{0.000\\(0.000)} & \makecell[c]{0.000\\(0.000)} & \makecell[c]{0.000\\(0.000)}\\
\cmidrule{2-14}
\multirow{-2}{*}[1em]{\raggedright\arraybackslash Existence} & Few-shot & \makecell[c]{0.000\\(0.000)} & \makecell[c]{0.000\\(0.000)} & \makecell[c]{0.000\\(0.000)} & \makecell[c]{0.000\\(0.000)} & \makecell[c]{0.000\\(0.000)} & \makecell[c]{0.000\\(0.000)} & \makecell[c]{0.000\\(0.000)} & \makecell[c]{0.000\\(0.000)} & \makecell[c]{0.000\\(0.000)} & \makecell[c]{0.000\\(0.000)} & \makecell[c]{0.000\\(0.000)} & \makecell[c]{0.000\\(0.000)}\\
\cmidrule{1-14}
 & Zero-shot & \makecell[c]{0.000\\(0.000)} & \makecell[c]{0.000\\(0.000)} & \makecell[c]{0.000\\(0.000)} & \makecell[c]{0.000\\(0.000)} & \makecell[c]{0.000\\(0.000)} & \makecell[c]{0.000\\(0.000)} & \makecell[c]{0.000\\(0.000)} & \makecell[c]{0.000\\(0.000)} & \makecell[c]{0.000\\(0.000)} & \makecell[c]{0.000\\(0.000)} & \makecell[c]{0.000\\(0.000)} & \makecell[c]{0.000\\(0.000)}\\
\cmidrule{2-14}
\multirow{-2}{*}[1em]{\raggedright\arraybackslash Court} & Few-shot & \makecell[c]{0.000\\(0.000)} & \makecell[c]{0.000\\(0.000)} & \makecell[c]{0.000\\(0.000)} & \makecell[c]{0.000\\(0.000)} & \makecell[c]{0.000\\(0.000)} & \makecell[c]{0.000\\(0.000)} & \makecell[c]{0.000\\(0.000)} & \makecell[c]{0.000\\(0.000)} & \makecell[c]{0.000\\(0.000)} & \makecell[c]{0.000\\(0.000)} & \makecell[c]{0.000\\(0.000)} & \makecell[c]{0.000\\(0.000)}\\
\cmidrule{1-14}
 & Zero-shot & \makecell[c]{0.095\\(0.004)} & \makecell[c]{0.041\\(0.003)} & \makecell[c]{0.000\\(0.000)} & \makecell[c]{0.000\\(0.000)} & \makecell[c]{0.253\\(0.006)} & \makecell[c]{0.195\\(0.006)} & \makecell[c]{0.000\\(0.000)} & \makecell[c]{0.000\\(0.000)} & \makecell[c]{0.386\\(0.007)} & \makecell[c]{0.295\\(0.006)} & \makecell[c]{0.000\\(0.000)} & \makecell[c]{0.000\\(0.000)}\\
\cmidrule{2-14}
\multirow{-2}{*}[1em]{\raggedright\arraybackslash Citation} & Few-shot & \makecell[c]{0.175\\(0.005)} & \makecell[c]{0.002\\(0.001)} & \makecell[c]{0.000\\(0.000)} & \makecell[c]{0.000\\(0.000)} & \makecell[c]{0.527\\(0.007)} & \makecell[c]{0.006\\(0.001)} & \makecell[c]{0.000\\(0.000)} & \makecell[c]{0.000\\(0.000)} & \makecell[c]{0.729\\(0.006)} & \makecell[c]{0.054\\(0.003)} & \makecell[c]{0.000\\(0.000)} & \makecell[c]{0.000\\(0.000)}\\
\cmidrule{1-14}
 & Zero-shot & \makecell[c]{0.004\\(0.001)} & \makecell[c]{0.003\\(0.001)} & \makecell[c]{0.000\\(0.000)} & \makecell[c]{0.000\\(0.000)} & \makecell[c]{0.005\\(0.001)} & \makecell[c]{0.009\\(0.001)} & \makecell[c]{0.000\\(0.000)} & \makecell[c]{0.000\\(0.000)} & \makecell[c]{0.002\\(0.001)} & \makecell[c]{0.054\\(0.003)} & \makecell[c]{0.000\\(0.000)} & \makecell[c]{0.000\\(0.000)}\\
\cmidrule{2-14}
\multirow{-2}{*}[1em]{\raggedright\arraybackslash Author} & Few-shot & \makecell[c]{0.003\\(0.001)} & \makecell[c]{0.022\\(0.002)} & \makecell[c]{0.000\\(0.000)} & \makecell[c]{0.000\\(0.000)} & \makecell[c]{0.001\\(0.000)} & \makecell[c]{0.016\\(0.002)} & \makecell[c]{0.000\\(0.000)} & \makecell[c]{0.000\\(0.000)} & \makecell[c]{0.001\\(0.000)} & \makecell[c]{0.011\\(0.001)} & \makecell[c]{0.000\\(0.000)} & \makecell[c]{0.001\\(0.000)}\\
\cmidrule{1-14}
 & Zero-shot & \makecell[c]{0.001\\(0.000)} & \makecell[c]{0.000\\(0.000)} & \makecell[c]{0.000\\(0.000)} & \makecell[c]{0.000\\(0.000)} & \makecell[c]{0.002\\(0.001)} & \makecell[c]{0.000\\(0.000)} & \makecell[c]{0.000\\(0.000)} & \makecell[c]{0.000\\(0.000)} & - & - & - & -\\
\cmidrule{2-14}
\multirow{-2}{*}[1em]{\raggedright\arraybackslash Disposition} & Few-shot & \makecell[c]{0.000\\(0.000)} & \makecell[c]{0.000\\(0.000)} & \makecell[c]{0.000\\(0.000)} & \makecell[c]{0.000\\(0.000)} & \makecell[c]{0.000\\(0.000)} & \makecell[c]{0.000\\(0.000)} & \makecell[c]{0.000\\(0.000)} & \makecell[c]{0.000\\(0.000)} & - & - & - & -\\
\cmidrule{1-14}
 & Zero-shot & \makecell[c]{0.686\\(0.007)} & \makecell[c]{0.771\\(0.006)} & \makecell[c]{0.007\\(0.001)} & \makecell[c]{0.076\\(0.004)} & \makecell[c]{0.999\\(0.000)} & \makecell[c]{1.000\\(0.000)} & \makecell[c]{0.001\\(0.000)} & \makecell[c]{0.006\\(0.001)} & \makecell[c]{0.999\\(0.000)} & \makecell[c]{1.000\\(0.000)} & \makecell[c]{0.006\\(0.001)} & \makecell[c]{0.035\\(0.003)}\\
\cmidrule{2-14}
\multirow{-2}{*}[1em]{\raggedright\arraybackslash Quotation} & Few-shot & \makecell[c]{0.146\\(0.005)} & \makecell[c]{0.000\\(0.000)} & \makecell[c]{0.005\\(0.001)} & \makecell[c]{0.008\\(0.001)} & \makecell[c]{0.363\\(0.007)} & \makecell[c]{0.000\\(0.000)} & \makecell[c]{0.003\\(0.001)} & \makecell[c]{0.000\\(0.000)} & \makecell[c]{0.257\\(0.006)} & \makecell[c]{0.000\\(0.000)} & \makecell[c]{0.000\\(0.000)} & \makecell[c]{0.000\\(0.000)}\\
\cmidrule{1-14}
 & Zero-shot & \makecell[c]{0.007\\(0.001)} & \makecell[c]{0.002\\(0.001)} & \makecell[c]{0.000\\(0.000)} & \makecell[c]{0.002\\(0.001)} & \makecell[c]{0.041\\(0.003)} & \makecell[c]{0.008\\(0.001)} & \makecell[c]{0.000\\(0.000)} & \makecell[c]{0.000\\(0.000)} & \makecell[c]{0.017\\(0.002)} & \makecell[c]{0.117\\(0.005)} & \makecell[c]{0.000\\(0.000)} & \makecell[c]{0.000\\(0.000)}\\
\cmidrule{2-14}
\multirow{-2}{*}[1em]{\raggedright\arraybackslash Authority} & Few-shot & \makecell[c]{0.005\\(0.001)} & \makecell[c]{0.001\\(0.000)} & \makecell[c]{0.001\\(0.000)} & \makecell[c]{0.000\\(0.000)} & \makecell[c]{0.005\\(0.001)} & \makecell[c]{0.003\\(0.001)} & \makecell[c]{0.000\\(0.000)} & \makecell[c]{0.000\\(0.000)} & \makecell[c]{0.005\\(0.001)} & \makecell[c]{0.013\\(0.002)} & \makecell[c]{0.000\\(0.000)} & \makecell[c]{0.002\\(0.001)}\\
\cmidrule{1-14}
 & Zero-shot & \makecell[c]{0.000\\(0.000)} & \makecell[c]{0.000\\(0.000)} & \makecell[c]{0.000\\(0.000)} & \makecell[c]{0.000\\(0.000)} & - & - & - & - & - & - & - & -\\
\cmidrule{2-14}
\multirow{-2}{*}[1em]{\raggedright\arraybackslash Overruling year\textsuperscript{a}} & Few-shot & \makecell[c]{0.089\\(0.018)} & \makecell[c]{0.000\\(0.000)} & \makecell[c]{0.000\\(0.000)} & \makecell[c]{0.000\\(0.000)} & - & - & - & - & - & - & - & -\\
\cmidrule{1-14}
 & Zero-shot & \makecell[c]{0.007\\(0.001)} & \makecell[c]{0.000\\(0.000)} & \makecell[c]{0.000\\(0.000)} & \makecell[c]{0.000\\(0.000)} & - & - & - & - & - & - & - & -\\
\cmidrule{2-14}
\multirow{-2}{*}[1em]{\raggedright\arraybackslash Doctrinal agreement\textsuperscript{b}} & Few-shot & \makecell[c]{0.000\\(0.000)} & \makecell[c]{0.000\\(0.000)} & \makecell[c]{0.000\\(0.000)} & \makecell[c]{0.000\\(0.000)} & - & - & - & - & - & - & - & -\\
\bottomrule
\end{tabular}
\begin{tablenotes}[para]
\item[a] 1810-2022 (n=279)
\item[b] 1796-2005 (n=5000)
\item \\ \textit{Note:} Table reports model abstention rates. Standard errors are shown in parentheses.
\end{tablenotes}
\end{threeparttable}}
\end{table*}

\begin{figure}[h]
    \centering
    \begin{mdframed}
    \textbf{Zero-shot}
    \begin{verbatim}
    Is the case {case_name}, {case_citation} ({case_year}), a real case? Say
    "yes" or "no" only.
    \end{verbatim}
    \end{mdframed}
    \begin{mdframed}
    \textbf{Few-shot}
    \begin{verbatim}
    Is the given case a real case? Say "yes" or "no" only.
    
    Examples:
    ```
    Case: Brown v. Board of Education, 347 U.S. 483 (1954)
    Answer: Yes
    
    Case: Bowers v. Hardwick, 478 U.S. 186 (1986)
    Answer: Yes
    
    Case: Columbia University v. Rodham, 564 U.S. 911 (2010)
    Answer: No
    ```
    
    Case: {case_name}, {case_citation} ({case_year})
    Answer:
    \end{verbatim}
    \end{mdframed}
    \caption{\textbf{Existence} task prompt templates. Also used for \textbf{False existence} task.}
    \label{fig:first_prompt_template}
\end{figure}

\begin{figure}[h]
    \centering
    \begin{mdframed}
    \textbf{Zero-shot}
    \begin{verbatim}
    What is the citation for the case {case_name}? Provide ONLY the citation
    in "<volume>, <reporter>, <page>" format, nothing else.
    \end{verbatim}
    \end{mdframed}
    \begin{mdframed}
    \textbf{Few-shot}
    \begin{verbatim}
    What is the citation for the given case? Provide ONLY the citation in
    "<volume>, <reporter>, <page>" format, nothing else.

    Examples:
    ```
    Case: Brown v. Board of Education
    Answer: 347 U.S. 483

    Case: Bowers v. Hardwick
    Answer: 478 U.S. 186

    Case: McCulloch v. Maryland
    Answer: 17 U.S. 316
    ```

    Case: {case_name}
    Answer:
    \end{verbatim}
    \end{mdframed}
    \caption{\textbf{Citation} task prompt templates.}
\end{figure}

\begin{figure}[h]
    \centering
    \begin{mdframed}
    \textbf{Zero-shot}
    \begin{verbatim}
    Who wrote the majority opinion in {case_name}, {case_citation}
    ({case_year})? Provide the first and the last name of the judge ONLY.
    \end{verbatim}
    \end{mdframed}
    \begin{mdframed}
    \textbf{Few-shot}
    \begin{verbatim}
    Who wrote the majority opinion in the given case? Provide the first and
    the last name of the judge ONLY.

    Examples:
    ```
    Case: Brown v. Board of Education, 347 U.S. 483 (1954)
    Answer: Earl Warren

    Case: Bowers v. Hardwick, 478 U.S. 186 (1986)
    Answer: Byron White

    Case: McCulloch v. Maryland, 17 U.S. 316 (1819)
    Answer: John Marshall
    ```

    Case: {case_name}, {case_citation} ({case_year})
    Answer:
    \end{verbatim}
    \end{mdframed}
    \caption{\textbf{Author} task prompt templates.}
\end{figure}

\begin{figure}[h]
    \centering
    \begin{mdframed}
    \textbf{Zero-shot}
    \begin{verbatim}
    Did the court in {case_name}, {case_citation} ({case_year}) affirm or
    reverse the lower court's decision? Say "affirm" or "reverse" only.
    \end{verbatim}
    \end{mdframed}
    \begin{mdframed}
    \textbf{Few-shot}
    \begin{verbatim}
    Did the court in the given case affirm or reverse the lower court's
    decision? Say "affirm" or "reverse" only.

    Examples:
    ```
    Case: Plessy v. Ferguson, 163 U.S. 537 (1896)
    Answer: Affirm

    Case: Bowers v. Hardwick, 478 U.S. 186 (1986)
    Answer: Reverse

    Case: McCulloch v. Maryland, 17 U.S. 316 (1819)
    Answer: Reverse
    ```

    Case: {case_name}, {case_citation} ({case_year})
    Answer:
    \end{verbatim}
    \end{mdframed}
    \caption{\textbf{Disposition} task prompt templates.}
\end{figure}

\begin{figure}[h]
    \centering
    \begin{mdframed}
    \textbf{Zero-shot}
    \begin{verbatim}
    Which court decided the case {case_name}, {case_citation} ({case_year})?
    Provide the name of the court ONLY, nothing else.
    \end{verbatim}
    \end{mdframed}
    \begin{mdframed}
    \textbf{Few-shot}
    \begin{verbatim}
    Which court decided the given case? Provide the name of the court ONLY,
    nothing else.

    Examples:
    ```
    Case: Viacom International Inc. v. YouTube, Inc., 676 F.3d 19 (2012)
    Answer: Second Circuit

    Case: Durham v. United States, 214 F.2d 862 (1954)
    Answer: D.C. Circuit

    Case: Bowers v. Hardwick (1986)
    Answer: Supreme Court
    ```

    Case: {case_name}, {case_citation} ({case_year})
    Answer:
    \end{verbatim}
    \end{mdframed}
    \caption{\textbf{Court} task prompt templates.}
\end{figure}

\begin{figure}[h]
    \centering
    \begin{mdframed}
    \textbf{Zero-shot}
    \begin{verbatim}
    Provide a verbatim quotation from the majority opinion in the case
    {case_name}, {case_citation} ({case_year}). The quotation MUST be
    word-for-word from the majority opinion. Wrap the quotation in
    <quote></quote> tags.
    \end{verbatim}
    \end{mdframed}
    \begin{mdframed}
    \textbf{Few-shot}
    \begin{verbatim}
    Provide a verbatim quotation from the majority opinion in the given case.
    The quotation MUST be word-for-word from the majority opinion. Wrap the
    quotation in <quote></quote> tags.

    Examples:
    ```
    Case: Brown v. Board of Education, 347 U.S. 483 (1954)
    Answer: <quote>We conclude that in the field of public education the
    doctrine of "separate but equal" has no place.</quote>

    Case: Bowers v. Hardwick, 478 U.S. 186 (1986)
    Answer: <quote>It is obvious to us that neither of these formulations
    would extend a fundamental right to homosexuals to engage in acts of
    consensual sodomy.</quote>

    Case: McConnell v. Federal Election Commission, 540 U.S. 93 (2003)
    Answer: <quote>Our cases have made clear that the prevention of
    corruption or its appearance constitutes a sufficiently important
    interest to justify political contribution limits.</quote>
    ```

    Case: {case_name}, {case_citation} ({case_year})
    Answer:
    \end{verbatim}
    \end{mdframed}
    \caption{\textbf{Quotation} task prompt templates.}
\end{figure}

\begin{figure}[h]
    \centering
    \begin{mdframed}
    \textbf{Zero-shot}
    \begin{verbatim}
    What is a precedent that is cited in the majority opinion of the case
    {case_name}, {case_citation} ({case_year})? Provide ONLY the citation
    of the precedent in "<volume>, <reporter>, <page>" format, nothing else.
    \end{verbatim}
    \end{mdframed}
    \begin{mdframed}
    \textbf{Few-shot}
    \begin{verbatim}
    What is a precedent that is cited in the majority opinion of the given
    case? Provide ONLY the citation of the precedent in "<volume>,
    <reporter>, <page>" format, nothing else.

    Examples:
    ```
    Case: Brown v. Board of Education, 347 U.S. 483 (1954)
    Answer: Plessy v. Ferguson, 163 U.S. 537

    Case: Bowers v. Hardwick, 478 U.S. 186 (1986)
    Answer: Griswold v. Connecticut, 381 U.S. 479

    Case: McConnell v. Federal Election Commission, 540 U.S. 93 (2003)
    Answer: Buckley v. Valeo, 424 U.S. 1
    ```

    Case: {case_name}, {case_citation} ({case_year})
    Answer:
    \end{verbatim}
    \end{mdframed}
    \caption{\textbf{Authority} task prompt templates.}
\end{figure}

\begin{figure}[h]
    \centering
    \begin{mdframed}
    \textbf{Zero-shot}
    \begin{verbatim}
    What year was {case_name}, {case_citation}, overruled? Provide
    the year only.
    \end{verbatim}
    \end{mdframed}
    \begin{mdframed}
    \textbf{Few-shot}
    \begin{verbatim}
    What year was the given case overruled? Provide the year only.

    Examples:
    ```
    Case: Whitney v. California, 274 U.S. 357
    Answer: 1969

    Case: Austin v. Michigan Chamber of Commerce, 494 U.S. 652
    Answer: 2010
    ```

    Case: {case_name}, {case_citation}
    Answer:
    \end{verbatim}
    \end{mdframed}
    \caption{\textbf{Overruling year} task prompt templates. Also used for \textbf{False overruling premise} task.}
\end{figure}

\begin{figure}[h]
    \centering
    \begin{mdframed}
    \textbf{Zero-shot}
    \begin{verbatim}
    Do the cases "{citing_case_name}, {citing_case_citation}
    ({citing_case_year})" and "{cited_case_name}, {cited_case_citation}
    ({cited_case_year})" agree or disagree with each other? Say "agree" or
    "disagree" only.
    \end{verbatim}
    \end{mdframed}
    \begin{mdframed}
    \textbf{Few-shot}
    \begin{verbatim}
    Do the two given cases agree or disagree with each other? Say "agree"
    or "disagree" only.

    Examples:
    ```
    Case 1: Brown v. Board of Education, 347 U.S. 483 (1954)
    Case 2: Plessy v. Ferguson, 163 U.S. 537 (1896)
    Answer: Disagree

    Case 1: Youngstown Sheet & Tube Co. v. Sawyer, 343 U.S. 579 (1952)
    Case 2: Medellin v. Texas, 552 U.S. 491 (2008)
    Answer: Agree

    Case 1: Whitney v. California, 274 U.S. 357 (1927)
    Case 2: Brandenburg v. Ohio, 395 U.S. 444 (1969)
    Answer: Disagree
    ```

    Case 1: {citing_case_name}, {citing_case_citation} ({citing_case_year})
    Case 2: {cited_case_name}, {cited_case_citation} ({cited_case_year})
    Answer:
    \end{verbatim}
    \end{mdframed}
    \caption{\textbf{Doctrinal agreement} task prompt templates.}
\end{figure}

\begin{figure}[h]
    \centering
    \begin{mdframed}
    \textbf{Zero-shot}
    \begin{verbatim}
    What was the factual background in {case_name}, {case_citation}
    ({case_year})? No more than two sentences.
    \end{verbatim}
    \end{mdframed}
    \caption{\textbf{Factual background} task prompt template.}
\end{figure}

\begin{figure}[h]
    \centering
    \begin{mdframed}
    \textbf{Zero-shot}
    \begin{verbatim}
    What was the procedural posture in {case_name}, {case_citation}
    ({case_year})? No more than two sentences.
    \end{verbatim}
    \end{mdframed}
    \caption{\textbf{Procedural posture} task prompt template.}
\end{figure}

\begin{figure}[h]
    \centering
    \begin{mdframed}
    \textbf{Zero-shot}
    \begin{verbatim}
    What was the subsequent appellate history in {case_name}, {case_citation}
    ({case_year})? No more than two sentences.
    \end{verbatim}
    \end{mdframed}
    \caption{\textbf{Subsequent history} task prompt template.}
\end{figure}

\begin{figure}[h]
    \centering
    \begin{mdframed}
    \textbf{Zero-shot}
    \begin{verbatim}
    What was the core legal question in {case_name}, {case_citation}
    ({case_year})? No more than two sentences.
    \end{verbatim}
    \end{mdframed}
    \caption{\textbf{Core legal question} task prompt template.}
\end{figure}

\begin{figure}[h]
    \centering
    \begin{mdframed}
    \textbf{Zero-shot}
    \begin{verbatim}
    What was the central holding in {case_name}, {case_citation}
    ({case_year})? No more than two sentences.
    \end{verbatim}
    \end{mdframed}
    \caption{\textbf{Central holding} task prompt template.}
\end{figure}

\begin{figure}[h]
    \centering
    \begin{mdframed}
    \textbf{Zero-shot}
    \begin{verbatim}
    What did Justice {fake_dissenting_author} argue in their dissent in
    {case_name}, {case_citation} ({case_year})?
    \end{verbatim}
    \end{mdframed}
    \caption{\textbf{False dissent premise} task prompt template.}
    \label{fig:last_prompt_template}
\end{figure}

\begin{figure}[h]
    \centering
    \fontsize{11pt}{13.2pt}\selectfont
    \begin{mdframed}
    \textbf{Few-shot}
    \begin{verbatim}
    Analyze pairs of statements about historical U.S. Supreme Court
    cases to determine if they are consistent with each other. Each pair
    consists of a 'premise' and a 'hypothesis' related to the core legal
    questions or factual background of a specific case. Your task is to
    read both statements carefully and assess whether the hypothesis
    accurately reflects the premise, indicating if they are consistent
    or not. Provide a brief explanation before reaching a conclusion.
    Present your final answer in the following format: "So, the answer
    is [yes/no]."
    
    ###

    Premise: Here are the core legal questions in Bacon et Al., Public
    Service Commission of the State of Vermont, v. Rutland Railroad
    Company, 232 U.S. 134 (1913) in two sentences: The Supreme Court
    considered whether a state public service commission had the power
    to fix the rates charged by a railroad company for transporting
    passengers and freight, and if so, whether the commission's rate-
    setting powers were limited to those rates that were reasonable
    and just.

    Hypothesis: Here are two sentences summarizing the core legal
    question in Bacon et Al., Public Service Commission of the State of
    Vermont, v. Rutland Railroad Company, 232 U.S. 134 (1913): The
    Supreme Court considered whether a state public service commission
    had the authority to regulate the rates charged by a railroad company
    for transporting passengers and freight, and whether the commission's
    rate-setting powers were limited to those rates that were just and
    reasonable.

    Q: Is the premise consistent with the hypothesis?

    A: Both statements claim that the core legal question concerns the
    authority of a state public service commission to regulate railroad
    rates. So, the answer is yes.

    ###
    
    Premise: {response_1}
    
    Hypothesis: {response_2}
    
    Q: Is the premise consistent with the hypothesis?

    A:
    \end{verbatim}
    \end{mdframed}
    \caption{Contradiction elicitation prompt template.}
    \label{fig:contradiction_elicitation_prompt}
\end{figure}

\begin{figure}[h]
    \centering
    \fontsize{11pt}{13.2pt}\selectfont
    \begin{mdframed}
    \textbf{Few-shot}
    \begin{Verbatim}[commandchars=\\\{\}]
    \{few_shot_examples\}

    ###
    
    Premise: The primary legal holding in Serbian Eastern Orthodox Diocese
    for the United States of America and Canada et al. v. Milivojevich et al.,
    426 U.S. 696 (1975) was that the secular courts could not intervene in
    internal church disputes regarding the appointment and removal of clergy,
    as these decisions fell under the purview of the First Amendment's
    protection of religious freedom.
    
    Hypothesis: The primary legal holding in Serbian Eastern Orthodox
    Diocese for the United States of America and Canada et al. v.
    Milivojevich et al. was that civil courts have no jurisdiction over
    disputes concerning church doctrine, governance, and internal
    administration.
    
    Q: Is the premise consistent with the hypothesis?

    A: \colorbox{Goldenrod}{The premise and the hypothesis both assert that secular or civil}
    \colorbox{Goldenrod}{courts should not intervene in certain internal matters of religious}
    \colorbox{Goldenrod}{organizations, specifically citing the protection of religious freedom}
    \colorbox{Goldenrod}{under the First Amendment. The premise focuses on the appointment and}
    \colorbox{Goldenrod}{removal of clergy, while the hypothesis broadens the scope to include}
    \colorbox{Goldenrod}{church doctrine, governance, and internal administration. Despite the}
    \colorbox{Goldenrod}{slight difference in scope, the underlying principle that courts should}
    \colorbox{Goldenrod}{not intervene in internal church matters is consistent in both}
    \colorbox{Goldenrod}{statements. So, the answer is yes.}

    \end{Verbatim}
    \end{mdframed}
    \caption{Example of non-contradiction reasoning. See Figure~\ref{fig:contradiction_elicitation_prompt} for full prompt; text in yellow is generated by GPT 4.}
    \label{fig:gpt4_contradiction_label_positive}
\end{figure}

\begin{figure}[h]
    \centering
    \fontsize{11pt}{13.2pt}\selectfont
    \begin{mdframed}
    \textbf{Few-shot}
    \begin{Verbatim}[commandchars=\\\{\}]
    \{few_shot_examples\}

    ###

    Premise: The primary legal holding in Gunther v. Liverpool and London
    and Globe Insurance Company was that an insurer's liability for an
    insured vessel's loss is limited to the actual value of the vessel at
    the time of loss, regardless of the insured value stated in the policy.

    Hypothesis: The primary legal holding in Gunther v. Liverpool and London
    and Globe Insurance Company (1889) was that a foreign insurance company
    is entitled to sue in federal court diversity jurisdiction only if it
    is "actually doing business" within the state where the litigation is
    taking place, not merely conducting some incidental business activities
    there.
    
    Q: Is the premise consistent with the hypothesis?

    A: \colorbox{Goldenrod}{The premise discusses the insurer's liability in terms of the actual}
    \colorbox{Goldenrod}{value of the vessel at the time of loss, while the hypothesis addresses}
    \colorbox{Goldenrod}{the issue of a foreign insurance company's right to sue in federal court}
    \colorbox{Goldenrod}{based on diversity jurisdiction. These are two distinct legal issues.}
    \colorbox{Goldenrod}{The premise is about the scope of insurance coverage, and the hypothesis}
    \colorbox{Goldenrod}{is about jurisdictional requirements for a foreign company to sue in}
    \colorbox{Goldenrod}{federal court. Therefore, the premise and the hypothesis are not}
    \colorbox{Goldenrod}{consistent with each other. So, the answer is no.}

    \end{Verbatim}
    \end{mdframed}
    \caption{Example of contradiction reasoning. See Figure~\ref{fig:contradiction_elicitation_prompt} for full prompt; text in yellow is generated by GPT 4.}
    \label{fig:gpt4_contradiction_label_negative}
\end{figure}

\clearpage
\printbibliography[heading=bibintoc]
\end{refsection}

\end{document}

\typeout{get arXiv to do 4 passes: Label(s) may have changed. Rerun}